\documentclass{article}

\usepackage[authoryear]{natbib}
 \usepackage[preprint]{neurips_2026}

\usepackage{multirow}
\usepackage[table]{xcolor}
\usepackage{hyperref}       % hyperlinks
\usepackage{url}            % simple URL typesetting
\usepackage{booktabs}       % professional-quality tables
\usepackage{amsfonts}       % blackboard math symbols
\usepackage{nicefrac}       % compact symbols for 1/2, etc.
\usepackage{microtype}      % microtypography
\usepackage{xcolor}         % colors
\usepackage{amsmath}
\usepackage{graphicx}
\usepackage{bm}
\usepackage{enumitem}
\usepackage{wrapfig}
\usepackage[ruled,vlined,linesnumbered]{algorithm2e}
\SetKwInput{KwRequire}{Require}
\newcommand{\eg}{\textit{e}.\textit{g}.}
\newcommand{\ie}{\textit{i}.\textit{e}.}
\newcommand{\cf}{\textit{cf.}}

\newcommand{\impr}[2]{#2$_{\textcolor{blue}{#1}}$}

\usepackage{color}
\definecolor{citecolor}{RGB}{66,168,235}
\definecolor{linkcolor}{RGB}{255,0,0}

\hypersetup{colorlinks=true,citecolor=citecolor,linkcolor=linkcolor}

\title{Direct Product Flow Matching: Decoupling Radial and Angular Dynamics for Few-Shot Adaptation}
\author{%
  Hongxu Chen\textsuperscript{1},
  Yanghao Wang\textsuperscript{1},
  Bowei Zhu\textsuperscript{2},
  Hongxiang Li\textsuperscript{1},
  \textbf{Zhen Wang}\textsuperscript{1}, \\
  \textbf{Ziqi Jiang}\textsuperscript{1},
  \textbf{Lin Li}\textsuperscript{1},
  \textbf{Rui Liu}\textsuperscript{3},
  \textbf{Long Chen}\textsuperscript{1}\thanks{Corresponding author.} \\
  \textsuperscript{1} HKUST, 
  \textsuperscript{2} USTC,
  \textsuperscript{3} Huawei Research
}

\begin{document}
\vspace{-2em}
\maketitle
\vspace{-1em}

\begin{abstract}

% Few-shot adaptation of vision-language models (VLMs) fundamentally relies on effective cross-modal alignment. Recent flow matching (FM) methods improve upon parameter-efficient fine-tuning (PEFT) by modeling alignment as a continuous multi-step unconditional probability flow. 
Recent flow matching (FM) methods improve the few-shot adaptation of vision-language models, by modeling cross-modal alignment as a continuous multi-step flow. In this paper, we argue that existing FM methods are inherently constrained by incompatible geometric priors on pre-trained cross-modal features, resulting in suboptimal adaptation performance. We first analyze these methods from a polar decomposition perspective (\ie, radial and angular sub-manifolds). Under this new geometric view, we identify three overlooked limitations in them: \textit{1) Angular dynamics distortion}: The radial-angular coupling induces non-uniform speed on the angular sub-manifold, leading to regression training difficulty and extra truncation errors. \textit{2) Radial dynamics neglect}: Feature normalization discards modality confidence, failing to distinguish out-of-distribution and in-distribution data, and abandoning crucial radial dynamics. \textit{3) Context-agnostic unconditional flow}: Dataset-specific information loss during pre-trained cross-modal feature extraction remains unrecovered. To resolve these issues, we propose \textbf{warped product flow matching (WP-FM)}, a unified Riemannian framework that reformulates alignment on a warped product manifold. Within this framework, we derive \textbf{direct product flow matching (DP-FM)} by introducing a constant-warping metric, which yields a decoupled cylindrical manifold (\ie, direct product manifold). DP-FM enables independent radial evolution and constant-speed angular geodesic transport, effectively eliminating angular dynamics distortion while preserving radial consistency. Meanwhile, we incorporate classifier-free guidance by conditioning the flow on the pre-trained VLMs' hidden states to inject missing dataset-specific information. Extensive results across 11 benchmarks have demonstrated that DP-FM achieves a new state-of-the-art for multi-step few-shot adaptation.
\end{abstract}

\section{Introduction}

The success of vision-language models (VLMs)~\citep{alayrac2022flamingo,li2023blip,liu2023visual,radford2021learning, li2022blip}, such as CLIP~\citep{radford2021learning}, fundamentally relies on cross-modal alignment. This capability embeds semantically corresponding visual and textual signals into a shared, expressive latent space, where their alignment is primarily characterized by \textit{angular similarity}. This precise alignment serves as the core mechanism enabling robust generalization and semantic understanding across diverse downstream tasks. However, when transferring these pre-trained VLMs to specific target domains with limited data, the pre-trained alignment often suffers from domain shift, requiring domain-specific fine-tuning (\ie, few-shot adaptation).

\begin{figure}
    \centering
    \vspace{-1em}
    \includegraphics[width=1.0\linewidth]{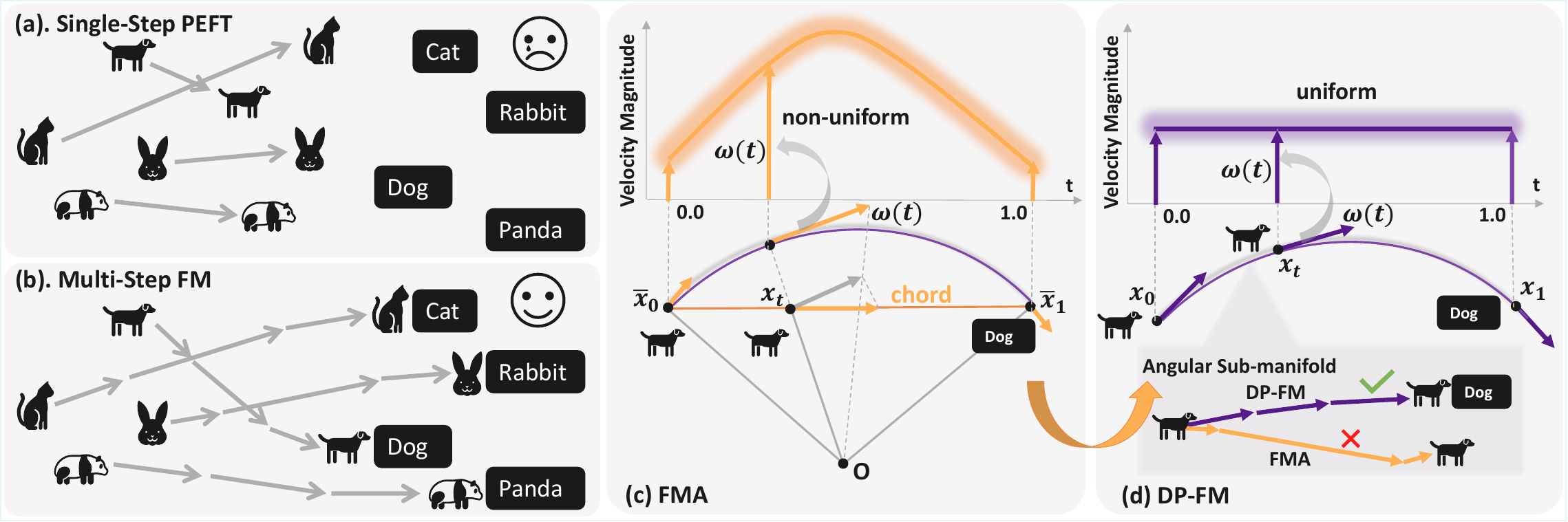}
    \caption{\textbf{(a).} Single-step parameter-efficient fine-tuning (PEFT) mostly performs cross-modal alignment in a single-step manner. \textbf{(b).} Multi-step flow matching (FM) methods model continuous and multi-step alignment dynamics. During the training stage, \textbf{(c).} \textcolor{orange}{FMA} undergoes a non-uniform angular speed induced by radial–angular coupling. However, \textbf{(d).} \textcolor{purple}{DP-FM} follows a constant-speed angular geodesic due to decoupled radial and angular dynamics.} 
    \label{fig:fig-1}
    \vspace{-1.5em}
\end{figure}

Traditionally, such adaptation relies on parameter-efficient fine-tuning (PEFT) methods~\citep{zhou2022conditional,zhou2022learning,zanella2024low,hu2022lora}, which can enhance the cross-modal alignment capability of the pretrained VLMs. However, PEFT methods typically perform alignment in a single-step adjustment, where cross-modal features align utilizing only one forward pass of the PEFT module (\cf, Figure~\ref{fig:fig-1}(a)), lacking an explicit mechanism to model \textit{continuous} and \textit{multi-step} alignment dynamics. As a result, they often struggle to resolve the alignment of highly entangled cross-modal distributions (\ie, image features and text features) in challenging datasets~\citep{jiang2025exploring}. To address this limitation, flow matching for few-shot adaptation (FMA)~\citep{jiang2025exploring} introduces a paradigm shift by conceptualizing cross-modal alignment as an \textit{unconditional}\footnote{An \textit{unconditional} flow indicates that the learned velocity field $v(\bm{x}_t, t)$ depends solely on the current state $\bm{x}_t$ and time $t$. It models the marginal probability path between two distributions without relying on any external conditions.} continuous probability flow.

\begin{wrapfigure}{r}{0.55\textwidth}
    \centering
    % 注意这里 width=\linewidth 是指图片占满这个 wrapfigure 区域的 100%
    \vspace{-2em}
    \includegraphics[width=\linewidth]{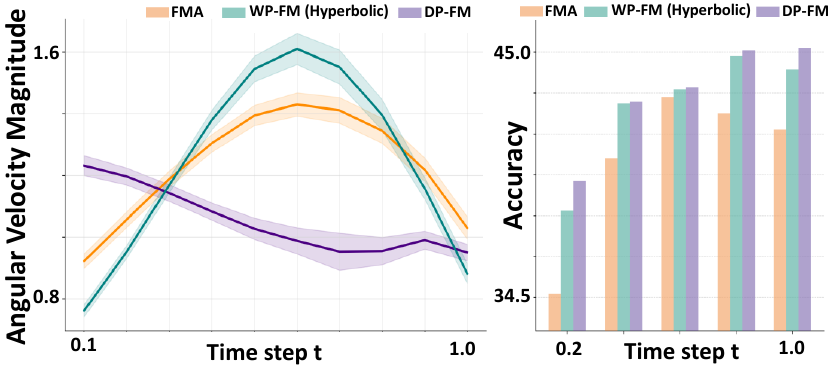}
    \vspace{-2em}
    \caption{\textbf{Comparison between FMA~\cite{jiang2025exploring}, WP-FM (Hyperbolic)~\cite{li2026path}, and DP-FM} on Aircraft dataset at 100 epochs. DP-FM shows enhanced accuracy and more uniform angular speed across time step $t$ at inference.} 
    \vspace{-0.5em}
    \label{fig:fig-2}
\end{wrapfigure}

 Specifically, FMA first normalizes the pre-trained cross-modal features (\eg, CLIP image and text features $\bm{x}_0, \bm{x}_1$) into unit-length pairs $(\bar{\bm{x}}_0, \bar{\bm{x}}_1)$, where $\bar{\bm{x}}_i = \bm{x}_i / \|\bm{x}_i\|_2$ for $i \in \{0, 1\}$. It then trains a velocity network to regress the target velocity $\bar{\bm{x}}_1 - \bar{\bm{x}}_0$ induced by the \textit{linear interpolation} trajectory $\bm{x}_t = (1-t)\bar{\bm{x}}_0 + t\bar{\bm{x}}_1$ on a flat manifold (\ie, the \textcolor{orange}{chord} as shown in Figure~\ref{fig:fig-1}(c)). By multi-step integration over the predicted velocity, FMA progressively transports the image features to the text features (Figure~\ref{fig:fig-1}(b)), thereby achieving improved alignment compared to previous PEFT methods.

To analyze the inherent mechanics of cross-modal alignment, it is natural to apply a \textbf{polar decomposition}~\citep{jo2026angular} to the interpolated features $\bm{x}_t$, disentangling them into a radial component ($r_t = \|\bm{x}_t\|_2$, \ie, radial sub-manifold $\mathcal{M}_r$, reflecting modality confidence) and an angular component ($\bm{\theta}_t = \bm{x}_t/\|\bm{x}_t\|_2$, \ie, angular sub-manifold $\mathcal{M}_{\bm{\theta}}$, encoding semantic direction). Under this geometric and probabilistic view, we argue that existing continuous flow matching (FM) methods~\citep{jiang2025exploring,li2026path} for few-shot adaptation remain bottlenecked, failing to exploit the underlying capabilities of FM with suboptimal adaptation performance mostly due to \textbf{three main limitations}:
\vspace{-0.8em}
\begin{enumerate}[label=\textbf{\arabic*)}, leftmargin=*]
\itemsep-0.2em
\item \textbf{Angular Dynamics Distortion.} \label{lim:limitation1} They impose an incompatible geometric prior that distorts the intrinsic alignment dynamics. In FMA, the linear interpolation trajectory on a flat manifold induces severe \textit{radial–angular coupling}. 
% As shown in Figure~\ref{fig:fig-1}(c), when the interpolated feature $\bm{x}_t$ traverses this \textcolor{orange}{chord}, the speed $\bm{\omega}(t)$\footnote{We use \textit{speed} as a scalar to denote the magnitude of velocity, and \textit{velocity} as a vector with both magnitude and direction.} on the angular sub-manifold (\ie, \textcolor{violet}{purple arc}) which is projected from velocity $\bar{\bm{x}}_1 - \bar{\bm{x}}_0$, undergoes angular dynamics distortion, characterized by \textit{non-uniform} speed with non-zero angular acceleration. 
As shown in Figure~\ref{fig:fig-1}(c), when the interpolated feature $\bm{x}_t$ traverses this \textcolor{orange}{chord}, the speed $\bm{\omega}(t)$\footnote{We use \textit{speed} as a scalar to denote the velocity magnitude, and \textit{velocity} represents both magnitude and direction.} on the angular sub-manifold (\ie, \textcolor{violet}{purple arc}) is projected from velocity $\bar{\bm{x}}_1 - \bar{\bm{x}}_0$. It undergoes angular dynamics distortion, characterized by \textit{non-uniform} speed with non-zero acceleration.
This induces extra curvature in the target velocity field, which increases the difficulty of regression training and leads to additional truncation errors, resulting in suboptimal performance during multi-step alignment.
The following work HFM~\citep{li2026path} implicitly bypasses this distortion by forcefully adjusting pre-trained features into a hyperbolic space~\citep{desai2023hyperbolic} before applying FM. However, it introduces unnecessary geometric complexity and destroys the inherent angular sub-manifold inherent to these pre-trained features, leading to incompatibility with standard, off-the-shelf adaptation pipelines widely used in current cross-modal tasks.
% \footnote{Throughout this paper, we use ``angular dynamics distortion'' to refer specifically to the induced non-uniform angular velocity caused by radial–angular coupling, which manifests as increased curvature in the velocity field and leads to additional truncation errors during multi-step alignment.}.

\item \textbf{Radial Dynamics Neglect.} \label{lim:limitation2} 
Radial information intrinsically reflects modality confidence, acting as an explicit signal to distinguish out-of-distribution (OOD) and in-distribution (ID) data~\citep{park2023understanding} (Refer to Appendix~\ref{app:modality_confidence}). This confidence is helpful for the velocity network in few-shot adaptation, as it enables the alignment process to remain sensitive to uncertain OOD samples and apply focused corrections. However, due to the inherent radial--angular coupling in existing methods, explicitly retaining this radial information exacerbates the distortion in angular dynamics, leading to degraded performance. 

% Consequently, current methods are forced to resort to normalization, 

% \ie, it directly abandons crucial radial information and dynamics, resulting in suboptimal performance.

\item \textbf{Context-Agnostic Unconditional Flow.} \label{lim:limitation3} 
In few-shot adaptation scenarios, cross-modal features extracted by pre-trained VLMs are highly generalized and lack dataset-specific details for the target domain. Existing FM methods model cross-modal alignment as an unconditional probability flow, relying solely on these generalized features. The problem is that an unconditional flow purely aligns pair-wise features, making it inherently \textit{context-agnostic}. Since it is blind to the whole dataset context, it cannot well adapt these generalized features into domain-specific ones. Therefore, there is no additional conditional guidance to introduce the target domain context into the alignment process, \ie, the adaptation performance inevitably degrades.
% \footnote{More experimental evidence are shown in the ablation studies.}
% In few-shot adaptation scenarios, pre-trained VLMs often discard dataset-specific information required for the target domain during cross-modal feature extraction. Existing FM methods model cross-modal alignment as an unconditional probability flow, relying solely on these pre-trained cross-modal features. Being inherently context-agnostic, the unconditional flow cannot recover this lost information. Without additional conditional guidance to inject the missing dataset-specific information back into the alignment process, the adaptation performance will degrade.
% \lc{this paragraph needs to polish}
\end{enumerate}
\vspace{-0.5em}

\noindent\textbf{Unified WP-FM Framework.} To systematically resolve the geometric Limitation~\ref{lim:limitation1} and ~\ref{lim:limitation2}, we propose \textbf{warped product flow matching (WP-FM)} framework, which elevates the aforementioned polar decomposition into the Riemannian framework~\citep{chen2023flow}. Rather than describing cross-modal alignment on a flat manifold, WP-FM structures the features as a warped product manifold $\mathcal{M} = \mathcal{M}_r \times_{g_\phi} \mathcal{M}_{\bm{\theta}}$. However, merely decoupling the features topologically does not resolve radial–angular coupling, as coupling behavior is dictated by the underlying Riemannian metric $g_{\phi}$. To capture this, we equip $\mathcal{M}$ with a unified metric tensor $g_{\phi}: ds^2 = dr^2 + \phi(r)^2 d{\bm{\theta}}^2$. Here, $ds$, $dr$, and $d{\bm{\theta}}$ denote the infinitesimal distance elements on the full manifold $\mathcal{M}$, the radial sub-manifold $\mathcal{M}_r$, and the angular sub-manifold $\mathcal{M}_{\bm{\theta}}$, respectively, while $\phi(r)$ acts as a \textit{radial warping function} that controls the coupling behavior. Under this general framework, we can define different metrics to instantiate different variants covering previous FM methods for adaptation:
\vspace{-0.5em}
\begin{itemize}
\itemsep-0.2em
    \item When $\phi(r)=r$, it degrades to FM on the flat manifold, as in FMA~\citep{jiang2025exploring}.
    \item When $\phi(r)=\text{sinh}(r)$, it degrades to FM on the Hyperbolic manifold, as in HFM~\citep{li2026path}.
\end{itemize}
\vspace{-0.5em}
It is worth noting that these non-constant warping functions (in above-mentioned variants) intrinsically couple the radial and angular dynamics, which are ``warped'', directly causing the distortion discussed previously. To ensure cross-modal features naturally evolve along dual, independent geodesics $\gamma(t) = (\gamma_r(t), \gamma_{\bm{\theta}}(t))$ with a constant-speed angular geodesic, we introduce \textbf{direct product flow matching (DP-FM)} under the unified WP-FM framework: by deriving the metric for unraveling the radial–angular coupling, we acquire the constant warping function (\ie, $\phi(r) = \text{CONSTANT}$), yielding FM on a decoupled cylindrical manifold (\ie, direct product manifold). In DP-FM, both Limitation~\ref{lim:limitation1} and ~\ref{lim:limitation2} are resolved: the radial trajectory $\gamma_r(t)$ preserves the dynamics of modality confidence, while the angular trajectory $\gamma_{\bm{\theta}}(t)$ independently traverses a \textit{constant-speed} geodesic (\cf, Figure~\ref{fig:fig-1}(d)), eliminating angular dynamics distortion for enhanced performance (\cf, Figure~\ref{fig:fig-2}). 

Furthermore, to resolve the context-agnostic Limitation~\ref{lim:limitation3} caused by information loss, DP-FM can incorporate classifier-free guidance (CFG)~\citep{ho2022classifier}. Instead of depending exclusively on the pre-trained generalized cross-modal features, we condition the flow on the pre-trained model's hidden states~\citep{rao2022denseclip}. This explicitly injects the missing dataset-specific information into the alignment, allowing DP-FM to recover fine-grained details in the target domain and guarantee enhanced few-shot adaptation.

Extensive experiments across 11 diverse few-shot benchmarks demonstrate that DP-FM achieves enhanced adaptation accuracy compared to previous state-of-the-art FM methods for few-shot adaptation.  In summary, our main contributions are threefold:
\textbf{1). Theoretical Insight:} We identify fundamental geometric limitations in existing continuous FM methods for few-shot adaptation, namely angular dynamics distortion and radial dynamics neglect.
\textbf{2). Unified WP-FM Framework:} We propose warped product FM to reformulate cross-modal alignment on a unified Riemannian manifold.
\textbf{3). DP-FM with CFG:} We derive DP-FM to strictly decouple radial and angular dynamics. Additionally, we integrate CFG to inject dataset-specific information back into the alignment process.
% \textbf{2). Unified Geometric \& Conditional Framework:} We introduce Warped Product Flow Matching (WP-FM), \textcolor{red}{which reformulates cross-modal alignment on a unified Riemannian manifold to strictly decouple radial and angular dynamics via an optimal metric in Direct Product Flow Matching (DP-FM)}. Furthermore, we equip the framework with CFG to explicitly inject dataset-specific information. 

% \textbf{3). xxx}
% \textbf{3). Empirical Results:} Extensive experiments demonstrate enhanced performance across multi-step alignment across 11 benchmarks.

\section{Related Work}

\noindent\textbf{Few-shot Adaptation in VLMs.}
Few-shot adaptation of vision-language models (VLMs)~\citep{radford2021learning,zhai2023sigmoid,jia2021scaling} aims to transfer pre-trained cross-modal alignment to downstream domains using limited data. To achieve this, prior work predominantly adopts parameter-efficient fine-tuning (PEFT) methods~\citep{han2024parameter}, which adapt models without updating the entire model. Among these approaches, prompt tuning methods~\citep{zhou2022learning, zhou2022conditional} optimize learnable context vectors, typically within the text encoder, while adapter-based methods~\citep{houlsby2019parameter,gao2024clip,hu2022lora} introduce lightweight trainable modules into the model. These methods have demonstrated effective empirical performance in few-shot adaptation. Recent works~\citep{jiang2025exploring, li2026path} explore continuous formulations based on flow matching, where cross-modal alignment is modeled as a multi-step unconditional probability flow. While they provide an expressive multi-step framework for cross-modal alignment, they fail to exploit the underlying capabilities of FM with suboptimal adaptation performance. As analyzed in Limitation~\ref {lim:limitation1} and ~\ref{lim:limitation2}, these issues are closely related to the underlying geometric limitations inherent in existing methods.

% PEFT methods generally operate in a single-step manner, implicitly performing cross-modal alignment in single-step adjustment, with one forward pass of the PEFT module. As a result, they do not explicitly model the continuous multi-step alignment dynamics, which can be highly entangled cross-modal distributions in challenging datasets. To address this limitation,
% Recent works~\citep{jiang2025exploring, li2026path} explore continuous formulations based on flow matching (FM), where cross-modal alignment is modeled as a multi-step unconditional probability flow. While they provide an expressive multi-step framework for cross-modal alignment, they fail to exploit the underlying capabilities of FM with suboptimal adaptation performance. As analyzed in Limitation~\ref {lim:limitation1} and ~\ref{lim:limitation2}, these issues are closely related to the underlying geometric limitations inherent in existing methods.

\noindent\textbf{Flow Matching (FM) and Riemannian FM.}
FM~\citep{lipman2022flow,liu2022flow} has emerged as an efficient framework for generative modeling~\citep{rombach2022high,ma2024sit,peebles2023scalable,chen2026bi}. By directly regressing target velocity fields induced by simple and tractable probability paths, FM enables a streamlined training pipeline and improved numerical stability compared to prior approaches (\eg, traditional diffusion models~\citep{ho2020denoising,song2020score}). To accommodate non-Euclidean data, Riemannian FM (RFM)~\citep{chen2023flow} constructs optimal transport trajectories under the Riemannian metric. While theoretically principled, RFM lacks a task-specific metric design. Recent FM method for few-shot adaptation~\citep{jiang2025exploring} defaults to incompatible geometries, ignoring the intrinsic radial and angular structure of pre-trained cross-modal features. To bridge this gap, our proposed warped product FM (WP-FM) tailors RFM for cross-modal alignment by introducing a unified metric with a radial warping function. Crucially, by adopting an optimal constant-warping metric, we derive direct product FM (DP-FM), which decouples radial and angular dynamics. This efficiently resolves Limitations~\ref{lim:limitation1} and \ref{lim:limitation2}, establishing a robust geometric foundation for multi-step adaptation.

\section{Methodology}
\subsection{Unified Warped Product Flow Matching (WP-FM) Framework}
\noindent\textbf{Preliminaries: Warped Product Manifolds.} In differential geometry, a \textit{warped product manifold}~\citep{o1983semi} provides a generalized structure for the product of two Riemannian manifolds. Given a base manifold $(\mathcal{B}, g_\mathcal{B})$ and a fiber manifold $(\mathcal{F}, g_\mathcal{F})$, their warped product manifold $\mathcal{M} = \mathcal{B} \times_f \mathcal{F}$ is constructed by equipping the product topological space $\mathcal{B} \times \mathcal{F}$ with a metric tensor characterized by the following line element:
$
    ds^2 = ds_\mathcal{B}^2 + f(x)^2 ds_\mathcal{F}^2,
    \label{eq:warped_metric}
$
where $f: \mathcal{B} \to \mathbb{R}_{>0}$ is a smooth scalar function called the \textit{warping function}. Under this structure, the intrinsic geometry of the fiber manifold $\mathcal{F}$ is modified (\ie, ``warped'') by $f(x)$, depending solely on the point $x \in \mathcal{B}$. This framework provides a foundation for pre-trained cross-modal feature representation, allowing us to map the modality confidence to the base manifold and the semantic direction to the fiber manifold.

\noindent\textbf{Riemannian FM on Warped Product Manifolds.} We begin by formalizing cross-modal alignment as a continuous probability flow on a Riemannian manifold. Given features $\bm{x} \in \mathbb{R}^d$ (\ie, any possible interpolated features between cross-modal features) in the cross-modal alignment, we apply a polar decomposition to disentangle the feature:
$
\bm{x} = r \cdot \bm{\theta},  r = \|\bm{x}\|_2,  \bm{\theta} \in \mathbb{S}^{d-1}.
\label{eq1}
$
This induces a warped product manifold $\mathcal{M} = \mathcal{M}_r \times_{g_\phi} \mathcal{M}_{\bm{\theta}}$, where $\mathcal{M}_r = \mathbb{R}_{>0}$ can be interpreted as reflecting modality confidence (\ie, base radial sub-manifold), and $\mathcal{M}_{\bm{\theta}} = \mathbb{S}^{d-1}$ encodes the semantic direction (\ie, fiber angular sub-manifold). To characterize the dynamics and radial-angular coupling behavior, we establish a unified Riemannian manifold $(\mathcal{M}, g_\phi)$, where the metric tensor $g_{\phi}$ is defined by a family of rotationally symmetric line elements:
$
ds^2 = dr^2 + \phi(r)^2 d\bm{\theta}^2,
\label{eq2}
$
where $ds$, $dr$, and $d{\bm{\theta}}$ denote the infinitesimal distance elements on the full manifold $\mathcal{M}$, the radial sub-manifold $\mathcal{M}_r$, and the angular sub-manifold $\mathcal{M}_{\bm{\theta}}$, respectively, $\phi(r)$ is a radial warping function. 

\noindent\textbf{Warped Product Flow Matching.} This formulation provides a unified geometric framework that subsumes a broad class of FM methods based on different choices of $g_{\phi}$ (\eg, $\phi(r)=r$ degrades to the flat manifold in FMA~\citep{jiang2025exploring}, $\phi(r)=\text{sinh}(r)$ degrades to the Hyperbolic manifold in HFM~\citep{li2026path}).
We define Riemannian FM on this warped product manifold as \textbf{warped product flow matching (WP-FM)}. Given paired pretrained cross-modal features $(\bm{x}_0, \bm{x}_1)$ (\ie, $\bm{x}_0$ is image feature, and $\bm{x}_1$ is text feature), the optimal transport trajectory corresponds to a geodesic $\gamma(t)$ under the chosen metric $g_{\phi}$. The target velocity fields regressed by the network are then defined as $\bm{v}^*(\bm{x}_t, t) = \frac{d}{dt} \gamma(t)$.  

\subsection{Analyses: Angular Dynamics Distortion}

% FMA~\cite{jiang2025exploring} operates as a suboptimal instantiation of WP-FM, along with discarding radial dynamics utilizing $L_2$ normalization to raw cross-modal features. It implicitly chooses an incompatible warping function $\phi(r)$ that inevitably introduces \textit{radial–angular coupling}, inducing \textit{angular dynamics distortion}. Specifically, setting $\phi(r) = r$ degrades the generalized metric to the standard Euclidean space in FMA. Under this geometric prior, the optimal transport geodesic defaults to a rigid linear interpolation: $\gamma(t) = (1 - t)\bar{\bm{x}}_0 + t \bar{\bm{x}}_1$, where $\bar{\bm{x}}_i = \bm{x}_i / \|\bm{x}_i\|_2$ for $i \in \{0, 1\}$.
FMA~\cite{jiang2025exploring} acts as a suboptimal instantiation of WP-FM that explicitly discards radial dynamics via $L_2$ normalization. By implicitly setting the warping function to $\phi(r) = r$, FMA degrades the metric to a flat space. Under this incompatible geometric prior, the optimal transport geodesic defaults to a linear interpolation: $\gamma(t) = (1 - t)\bar{\bm{x}}_0 + t \bar{\bm{x}}_1$, where $\bar{\bm{x}}_i = \bm{x}_i / \|\bm{x}_i\|_2$. While this trajectory maintains a constant velocity in the ambient space, it inherently introduces \textit{radial–angular coupling} and induces severe \textit{angular dynamics distortion}. Specifically, let ${\gamma}_r(t) = \|\gamma(t)\|_2$ denote the radial magnitude and $\omega(t) = \|\dot{\gamma}_{\bm{\theta}}(t)\|_2$ denote the angular speed, where ${\gamma}_{\bm{\theta}}(t) := \gamma(t) / \|\gamma(t)\|_2$. Under this interpolation, the trajectory behaves as a linear chord, causing the magnitude ${{\gamma}_r}(t)$ to sag ($\dot{{\gamma}_r}(t) \neq 0$). This geometric sagging amplifies the angular speed near the midpoint ($t=0.5$). Taking the time derivative of the angular speed yields a non-zero variation in angular velocity:
\begin{equation}
\dot{\omega}(t) \propto - {\dot{{\gamma}_r}(t)}/{\gamma_r(t)^3} \neq 0.
\label{eq3}
\end{equation}
Eq.~\eqref{eq3} shows that the angular dynamics are actually affected by radial dimension in the flat space (\ie, radial–angular coupling). This incompatible geometry forces the trajectory to exhibit \textit{non-uniform angular speed} with non-zero angular acceleration (\cf, Figure~\ref{fig:fig-1}(c)). It inherently introduces additional curvature into the effective velocity field on the angular sub-manifold, exacerbating the difficulty of regression training and leading to extra truncation errors\footnote{ Derivations are in the Appendix~\ref{app:rigorous_proof}, showing that any non-constant warping metric induces an extra truncation error.}, which causes suboptimal alignment performance during multi-step alignment for adaptation. If FMA retains the radial information without any normalization, the angular speed will be even more non-uniform.

The following work HFM~\citep{li2026path} forcefully adjusts the pre-trained cross-modal features into a hyperbolic space~\citep{desai2023hyperbolic} before applying FM via centripetal hyperbolic alignment, which can implicitly bypass this distortion. Although adjusting features toward a hyperbolic boundary can artificially boost class separability to achieve competitive few-shot performance, this extra mapping step introduces an inefficient two-stage training paradigm and unnecessary geometric complexity, which disrupts the inherent angular sub-manifold of pre-trained cross-modal features, leading to incompatibility with standard, off-the-shelf adaptation pipelines widely used in current cross-modal tasks.

\subsection{Direct Product Flow Matching (DP-FM)}

To eliminate this distortion while preserving the inherent manifold of pre-trained cross-modal representations, the angular trajectory should independently follow a \textit{constant-speed geodesic} on $\mathbb{S}^{d-1}$. In Riemannian geometry, this requires the covariant derivative of the angular velocity to strictly vanish: $\frac{D}{dt}\dot{{\gamma}}_{\bm{\theta}}(t) = \mathbf{0}$. Under the generalized metric, expanding the angular acceleration via Christoffel symbols~\citep{o1983semi} ($\Gamma^\theta_{r\theta} = \frac{\phi'(r)}{\phi(r)}$) reveal the coupling:
\begin{equation}
\frac{D}{dt}\dot{{\gamma}}_{\bm{\theta}}(t) = -2 \frac{\phi'(\gamma_r(t))}{\phi(\gamma_r(t))} \dot{\gamma}_r(t) \dot{\gamma}_{\bm{\theta}}(t).
\label{eq4}
\end{equation}
% For the angular trajectory to maintain a constant speed, the coupling term must be universally eliminated. 
This mandates $\phi'(\gamma_r(t)) = 0$, leading to the metric:
$
\phi(r) = \text{CONSTANT}.
\label{eq5}
$
This constant-warping metric establishes a theoretically optimal product manifold, \ie, a decoupled cylindrical manifold $\mathbb{R}_{>0} \times \mathbb{S}^{d-1}$. Driven by this finding, we propose \textbf{direct product flow matching (DP-FM)}. By constructing the continuous flow exclusively under this optimal metric, the target velocity field naturally decomposes into two independent dynamic objectives: $\bm{v}^*(\bm{x}_t, t) = \frac{d}{dt} \gamma(t)=( \dot{\gamma}_r(t), \dot{\gamma}_{\bm{\theta}}(t) )$. Specifically, given an image feature $\bm{x}_0 = r_0 \bm{\theta}_0$ and a text feature $\bm{x}_1 = r_1 \bm{\theta}_1$, the decoupled ground-truth geodesics are formulated as:
\begin{equation}
\gamma_r(t) = (1 - t)r_0 + t r_1, \quad\gamma_{\bm{\theta}}(t) = \frac{\sin((1-t)\alpha)}{\sin \alpha}\bm{\theta}_0 + \frac{\sin(t\alpha)}{\sin \alpha}\bm{\theta}_1,
\label{eq6}
\end{equation}
where $\alpha = \arccos(\langle \bm{\theta}_0, \bm{\theta}_1 \rangle) \in [0, \pi]$.
This explicit dual-geodesic dynamics cleanly resolves the two geometric limitations of previous methods: \textbf{1). Mitigating Angular Dynamics Distortion:} As formulated in Eq.~\eqref{eq6}, the angular geodesic $\gamma_{\bm{\theta}}(t)$ naturally evolves via exact spherical linear interpolation (slerp) on $\mathbb{S}^{d-1}$. This guarantees a constant-speed geodesic ($\frac{D}{dt}\dot{{\gamma}}_{\bm{\theta}}(t) = \mathbf{0}$), fundamentally eliminating the angular dynamics distortion, thereby preventing extra curvature in the target velocity field. \textbf{2). Addressing Radial Dynamics Neglect:} Rather than discarding radial information, the radial geodesic $\gamma_r(t)$ is explicitly retained and modeled as an independent radial trajectory, which preserves modality confidence to distinguish OOD from ID samples for the velocity network.

\begin{algorithm*}[tb]
\caption{Warped Product Flow Matching (WP-FM)}\label{alg:pfm}
\vspace{1mm}
\textbf{Phase 1: Training} \\
\KwRequire{Image and text features $\mathcal{P}_0$, $\mathcal{P}_1$, velocity field $\bm{v}_\psi$, steps $N$, dropout $p_{\text{drop}}$.}

\While{not converged}{
    % [UPDATE] Added condition c to sampling
    Sample $\bm{x}_0 \sim \mathcal{P}_0, \bm{x}_1 \sim \mathcal{P}_1, t \sim \mathcal{U}(0, 1)$, extract hidden states condition $\bm{c}$\;
    Time-shift schedule $t \leftarrow \frac{s \cdot t}{1 + (s - 1)t}$, set $\bm{c} \leftarrow \emptyset$ with probability $p_{\text{drop}}$\;
    Polar decompose $r_i \leftarrow \|\bm{x}_i\|_2, \bm{\theta}_i \leftarrow \bm{x}_i / r_i$ ($i \in \{0, 1\}$)\;
    Compute coupled geodesics $(\gamma_r(t), \gamma_{\bm{\theta}}(t))$ and target velocities $(\dot{\gamma}_r(t), \dot{\gamma}_{\bm{\theta}}(t))$\;
    % Governed by generalized metric $g_{\phi}$
    % [UPDATE] Pass condition c to velocity prediction
    Assemble feature $\bm{x}_t \leftarrow \gamma_r(t) \cdot \gamma_{\bm{\theta}}(t)$, predict velocity $\bm{v}_\psi \leftarrow \bm{v}_\psi(\bm{x}_t, t| \bm{c})$\;
    Project to tangent spaces: $v_{rad} \leftarrow \langle \bm{v}_\psi, \gamma_{\bm{\theta}}(t) \rangle$, $\bm{v}_{ang} \leftarrow ({\bm{v}_\psi -v_{rad}\gamma_{\bm{\theta}}(t)})/{\gamma_r(t)}$\;
    Compute metric-aware loss: $\mathcal{L} \leftarrow \|v_{rad} -\dot{\gamma}_r(t)\|_2^2 + \phi(\gamma_r(t))^2 \|\bm{v}_{ang} - \dot{\gamma}_{\bm{\theta}}(t)\|_2^2$\;
    Update network parameters $\psi$ via $\nabla_\psi \mathcal{L}$\;
}

\vspace{2mm}
\hrule
\vspace{2mm}

\textbf{Phase 2: Inference} \\
% [UPDATE] Added target condition c
\KwRequire{Image feature $\bm{x}_{test} \sim \mathcal{P}_0$, hidden states condition $\bm{c}$, CFG scale $w$.}
Initialize $r_0 \leftarrow \|\bm{x}_{test}\|_2, \bm{\theta}_0 \leftarrow \bm{x}_{test} / r_0$, set step size $\Delta t \leftarrow 1/N$\;
\For{$k = 0$ \KwTo $N-1$}{
    State $\bm{x}_k \leftarrow r_k \cdot \bm{\theta}_k$ at time $t \leftarrow k \Delta t$\;
    % [UPDATE] Replaced simple prediction with CFG guided velocity extrapolation
    Predict condition-guided velocity: $\tilde{\bm{v}}_\psi \leftarrow \bm{v}_\psi(\bm{x}_k, t| \emptyset) + w \cdot \big(\bm{v}_\psi(\bm{x}_k, t| \bm{c}) - \bm{v}_\psi(\bm{x}_k, t| \emptyset)\big)$\;
    % [UPDATE] Use the guided velocity \tilde{\bm{v}}_\psi for tangent projection
    Project to tangent spaces: $v_{rad} \leftarrow \langle \tilde{\bm{v}}_\psi, \bm{\theta}_k \rangle$, $\bm{v}_{ang} \leftarrow ({\tilde{\bm{v}}_\psi - v_{rad}\bm{\theta}_k})/{r_k}$\;
    Integration on the manifold: $(r_{k+1}, \bm{\theta}_{k+1}) \leftarrow \text{Exp}_{(r_k, \bm{\theta}_k)}^{\mathcal{M}}([v_{rad}, \bm{v}_{ang}] \Delta t)$\;
}
\KwOut{$\bm{x}_N \leftarrow r_N \cdot \bm{\theta}_N$\;}
\end{algorithm*}

\subsection{Algorithmic Design \& Implementation of WP-FM}

We introduce several critical designs for both the training and inference stages for WP-FM. The detailed algorithm is shown in Algorithm~\ref{alg:pfm}.

\textbf{Classifier-Free Guidance in Adaptation.} As identified in Limitation~\ref{lim:limitation3}, the pre-trained VLMs often yield generalized features that discard dataset-specific information in few-shot scenarios. To explicitly inject the missing information back into the alignment process, we equip WP-FM with classifier-free guidance (CFG)~\citep{ho2022classifier}, where the conditioning signal $\bm{c}$ is derived from the pre-trained model's hidden states given the input $\bm{x}_0$. During training, we parameterize conditional vector fields $v_\theta(\bm{x}_t, t|\bm{c})$. We jointly train an unconditional field by randomly replacing $\bm{c}$ with a learnable null token $\emptyset$ at a fixed dropout probability $p_{\text{drop}}$. During inference, we construct a semantic-guided velocity field $\tilde{v}_\psi$:
\begin{equation}
\tilde{v}_\psi(\bm{x}_t, t| \bm{c}) = v_\psi(\bm{x}_t, t| \emptyset) + w \cdot \Big( v_\psi(\bm{x}_t, t| \bm{c}) - v_\psi(\bm{x}_t, t| \emptyset) \Big),
\label{eqCFG}
\end{equation}
where $w \geq 1$ is the guidance scale. By integrating over $\tilde{v}_\psi$, DP-FM amplifies the target-domain information. This effectively steers the flow away from context-agnostic trajectories, recovering the fine-grained information discarded during pre-trained feature extraction for enhanced adaptation.

\begin{wrapfigure}{r}{0.45\textwidth} 
    \centering
    % 注意这里 width=\linewidth 是指图片占满这个 wrapfigure 区域的 100%
    \vspace{-1em}
    \includegraphics[width=\linewidth]{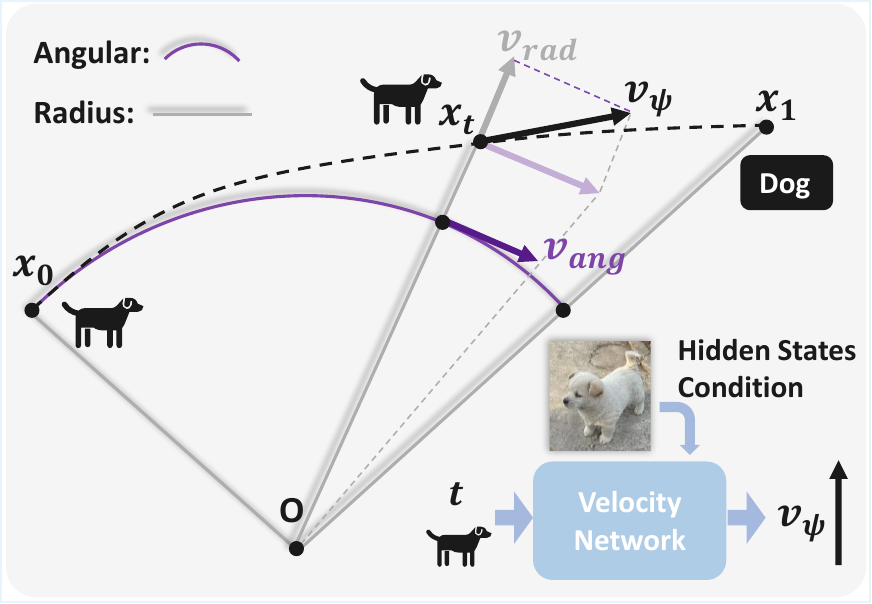}
    \vspace{-2em}
    \caption{\textbf{Illustration of DP-FM.}} 
    \vspace{-1em}
    \label{fig:fig-3}
\end{wrapfigure}
\textbf{Velocity Decomposition.}
Standard neural networks typically output velocity predictions in the ambient Euclidean space, denoted as $\bm{v}_\psi(\bm{x}_t, t) \in \mathbb{R}^d$. To adhere to our warped product manifold geometry, this output is explicitly projected onto the tangent spaces $\mathcal{T}\mathcal{M}_r$ and $\mathcal{T}\mathcal{M}_{\bm{\theta}}$. Specifically, as shown in Figure~\ref{fig:fig-3}, the radial velocity is extracted via the inner product: $v_{rad} = \langle \bm{v}_\psi, \gamma_{\bm{\theta}}(t) \rangle$. For the angular component, we isolate the orthogonal projection and fundamentally divide it by the radius $r_t$:
\begin{equation}
\bm{v}_{ang} = \frac{1}{\gamma_r(t)}\left({\bm{v}_\psi - v_{rad}\gamma_{\bm{\theta}}(t)} \right),
\label{eq7}
\end{equation}
which converts the predicted velocity in ambient Euclidean space into the angular velocity measured in radians, ensuring strict dimensionality alignment with the ground-truth geodesic.

\textbf{Metric-Aware Loss Objective.}
Unlike standard FM, which minimizes mean squared error (MSE) uniformly across Euclidean dimensions, WP-FM utilizes a specific optimization objective function corresponding to the Riemannian manifold. Guided by the metric tensor $ds^2 = dr^2 + \phi(r)^2 d\bm{\theta}^2$, the regression loss for WP-FM is formulated as a metric-aware objective:
\begin{equation}
\vspace{-0.5em}
\mathcal{L}_{\text{WP-FM}} = \mathbb{E}_{t, \bm{x}_0, \bm{x}_1} \left[ \|v_{rad} - \dot{\gamma}_r(t)\|^2_2 + \phi(\gamma_r(t))^2 \|\bm{v}_{ang} - \dot{\gamma}_{\bm{\theta}}(t)\|^2_2 \right].
\label{eq8}
\end{equation}
This ensures that the gradient updates strictly follow the geometric structure of the manifold.

\textbf{Inference via Exponential Map.}
During the inference phase\footnote{In the inference stage, we utilize $r_t$ and $\bm{\theta}_t$ to replace $\gamma_r(t)$ and $\gamma_{\bm{\theta}}(t)$.} of WP-FM, discrete multi-step integration (\eg, Euler method) in ambient space easily drifts off the manifold, leading to severe error accumulation. To guarantee mathematical closure, our method utilizes the formal Riemannian exponential map $\text{Exp}_{\bm{x}}^{\mathcal{M}}$~\citep{chen2023flow} for the ordinary differential equation (ODE) integration step $\Delta t$:
\begin{equation}
(r_{t+\Delta t}, \bm{\theta}_{t+\Delta t}) \leftarrow \text{Exp}_{(r_t, \bm{\theta}_t)}^{\mathcal{M}}([v_{rad}, \bm{v}_{ang}] \Delta t).
\label{eq9}
\end{equation}
\textbf{Time-shift Schedule.} Standard uniform time sampling allocates insufficient capacity to the complex initial phase of cross-modal alignment. To address this, we apply a non-linear time-shift schedule, which is widely used in high-resolution generative modeling~\citep{esser2024scaling, teng2023relay,hoogeboom2023simple}: $t' = \frac{s \cdot t}{1 + (s - 1)t}$, where $s$ is a time-shift parameter. Setting $s < 1$ monotonically biases the sampling density toward $t \to 0$, forcing the network to prioritize early velocity estimation and accelerating overall convergence.
\vspace{-0.5em}
\section{Experiments}

\begin{table*}[t]
    \centering
    \caption{\textbf{Comparison with Pure FM methods for few-shot adaptation.}}
    \label{tab:tab-1}
    \small
    \setlength{\tabcolsep}{3.5pt} % 压缩列间距防止超宽
    \resizebox{\textwidth}{!}{
    % 增加了一列 c，总共 15 列
    \begin{tabular}{@{} c| l| c c c c c >{\columncolor{gray!15}}l |cccccc >{\columncolor{gray!15}}l @{}}
        \toprule
        Shots & Methods &  Aircraft & SAT & DTD & SUN & Cars & \textbf{AVG} & Pets & UCF & Flowers & Caltech & Food & Net & \textbf{AVG} \\
        \midrule
        
        0 & CLIP~\citep{radford2021learning}  & 24.8 & 47.8 & 43.8 & 62.6 & 65.5 & 48.9 & 89.1 & 66.8 & 71.2 & 92.9 & 86.1 & 66.7 & 78.8 \\
        \midrule
        
        % 1-shot
        \multirow{3}{*}{1} 
        & WP-FM (Euclidean)~\citep{jiang2025exploring}   & 26.8 & 66.7 & 53.5 & 66.3 & 67.0 & \impr{+7.2}{56.1} & 90.2 & 71.2 & 82.2 & 93.9 & 86.1 & 67.5 & \impr{+3.1}{81.9} \\
        & WP-FM (Hyperbolic)  ~\citep{li2026path}  & 26.3 & 66.5 & 53.5 & 65.7 & 67.4 & \impr{+7.0}{55.9} & 89.8 & 71.3 & 81.7 & 94.1 & 86.1 & 67.5 & \impr{+3.0}{81.8} \\
        & DP-FM  & 26.3 & 68.6 & 53.3 & 66.3 & 67.2 & \impr{+7.4}{\textbf{56.3}} & 90.5 & 72.2 & 83.2 & 94.0 & 86.1 & 67.4 & \impr{+3.4}{\textbf{82.2}} \\
        \midrule
        
        % 4-shot
        \multirow{3}{*}{4} 
        & WP-FM (Euclidean)~\citep{jiang2025exploring}   & 30.7 & 77.9 & 61.4 & 69.9 & 71.0 & \impr{+13.3}{62.2} & 91.8 & 77.8 & 92.3 & 95.2 & 86.4 & 68.9 & \impr{+6.6}{85.4} \\
        & WP-FM (Hyperbolic)  ~\citep{li2026path}   & 32.4 & 78.5 & 61.1 & 70.1 & 72.6 & \impr{+14.0}{62.9} & 91.9 & 78.5 & 94.3 & 94.4 & 86.5 & 68.8 & \impr{+6.9}{85.7} \\
        & DP-FM  & 32.4 & 79.9 & 61.9 & 70.6 & 71.9 & \impr{+14.4}{\textbf{63.3}} & 92.1 & 78.1 & 93.6 & 95.5 & 86.5 & 69.0 & \impr{+7.0}{\textbf{85.8}} \\
        \midrule
        % 16-shot
        \multirow{3}{*}{16} 
        & WP-FM (Euclidean)~\citep{jiang2025exploring}   & 44.6 & 87.0 & 70.6 & 74.4 & 82.7 & \impr{+23.0}{71.9} & 92.9 & 82.7 & 98.3 & 95.7 & 86.8 & 71.1 & \impr{+9.1}{87.9} \\
        & WP-FM (Hyperbolic)  ~\citep{li2026path}   & 44.2 & 86.9 & 71.8 & 73.5 & 83.2 & \impr{+23.0}{71.9} & 93.1 & 83.9 & 97.8 & 95.8 & 86.9 & 71.0 & \impr{+9.3}{\textbf{88.1}} \\
        & DP-FM  & 46.0 & 87.8 & 71.2 & 73.8 & 83.6 & \impr{+23.6}{\textbf{72.5}} & 93.2 & 83.4 & 97.8 & 96.2 & 87.0 & 71.1 & \impr{+9.3}{\textbf{88.1}} \\
        \bottomrule
    \end{tabular}
    }
    \vspace{-1em}
\end{table*}

\begin{figure}[t]
    \centering
    \includegraphics[width=1\linewidth]{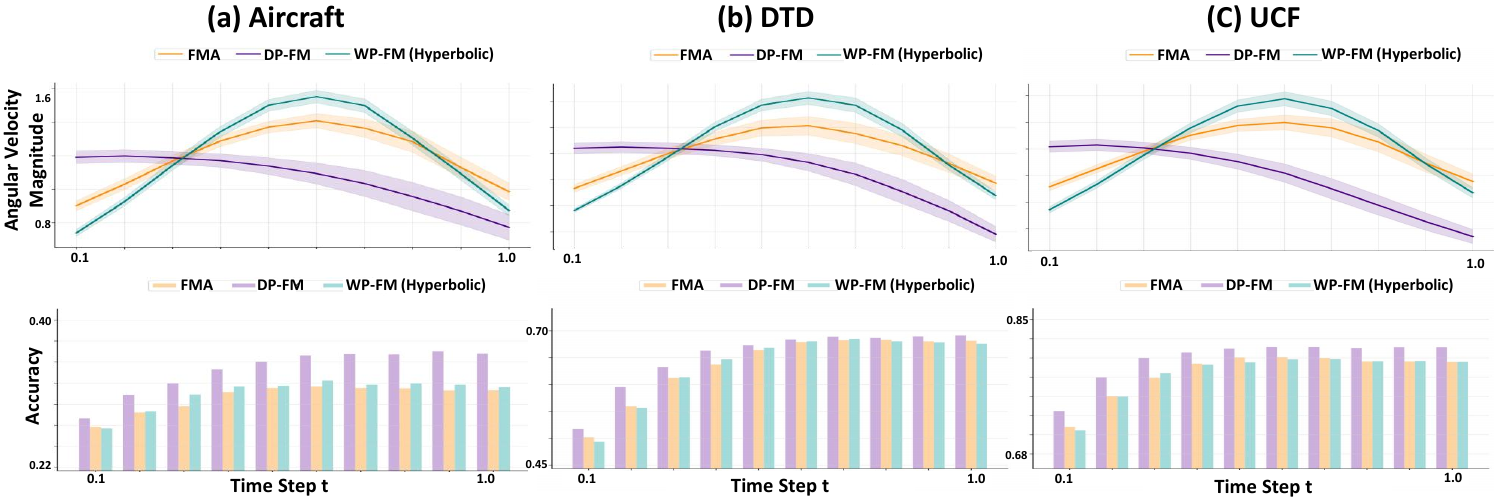}
    \vspace{-2em}
    \caption{\textbf{Comparison between WP-FM (Euclidean~\citep{jiang2025exploring}, Hyperbolic~\citep{li2026path}) and DP-FM} on (a). Aircraft, (b). DTD, and (c). UCF dataset at 20 epochs.} 
    \label{fig:fig-4}
    \vspace{-1.5em}
\end{figure}

\begin{table*}[t]
    \centering
    \caption{\textbf{Comparison with state-of-the-art FM methods for adaptation} building upon CLIP-LoRA and CLIP-Adapter. $^\dagger$ denotes that the PEFT module is trained using centripetal hyperbolic alignment~\citep{desai2023hyperbolic}. $\Delta$ represents the average performance improvement by FM over the baseline.}
    \label{tab:tab-2}
    \small
    \setlength{\tabcolsep}{3.5pt} 
    \resizebox{\textwidth}{!}{
    \begin{tabular}{@{} c| l| ccccc >{\columncolor{gray!15}}c >{\columncolor{gray!15}}c | cccccc >{\columncolor{gray!15}}c >{\columncolor{gray!15}}c @{}}
        \toprule
        & & \multicolumn{7}{c}{\textbf{Difficult Datasets}} & \multicolumn{8}{c}{\textbf{Easy Datasets}} \\
        \cmidrule(lr){3-9} \cmidrule(lr){10-17}
        Shot & Methods &  Aircraft & SAT & DTD & SUN & Cars & \textbf{AVG} & $\Delta$ & Pets & UCF & Flowers & Caltech & Food & Net & \textbf{AVG} & $\Delta$ \\
        \midrule
        
        \multirow{5}{*}{\textbf{1}} 
        & CLIP-LoRA$^\dagger$~\citep{desai2023hyperbolic} & 29.2 & 73.1 & 56.4 & 69.9 & 68.7 & 59.5 & - & 92.1 & 75.8 & 81.9 & 94.5 & 86.8 & 69.7 & 83.5 & - \\
        & \quad +HFM~\citep{li2026path}   & 30.5 & 76.8 & 57.4 & 71.2 & 69.8 & 61.1 & +1.7 & 92.2 & 76.6 & 85.8 & 95.1 & 85.9 & 70.0 & 84.3 & +0.8 \\
        & CLIP-LoRA~\citep{zanella2024low} & 28.0 & 71.9 & 54.1 & 70.3 & 69.4 & 58.7 & - & 91.9 & 75.4 & 81.4 & 93.8 & 85.1 & 70.3 & 83.0 & - \\
        & \quad +FMA~\citep{jiang2025exploring}   & 28.3 & 73.0 & 55.1 & 70.6 & 69.8 & 59.4 & +0.6 & 92.1 & 75.9 & 84.9 & 94.5 & 85.2 & 70.2 & 83.8 & +0.8 \\
        & \quad +DP-FM  & 29.6 & 81.7 & 56.9 & 71.0 & 70.8 & \textbf{62.0} & \textbf{+3.3} & 93.1 & 76.8 & 89.0 & 94.4 & 85.0 & 70.3 & \textbf{84.8} & \textbf{+1.8} \\
        \midrule

        \multirow{5}{*}{\textbf{4}} 
        & CLIP-LoRA$^\dagger$~\citep{desai2023hyperbolic} & 40.3 & 89.0 & 66.5 & 74.1 & 77.4 & 69.5 & - & 93.4 & 81.0 & 93.7 & 95.7 & 86.8 & 71.4 & 87.0 & - \\
        & \quad +HFM~\citep{li2026path}   & 43.8 & 90.7 & 67.6 & 75.5 & 79.9 & \textbf{71.5} & +2.0 & 93.7 & 83.4 & 95.5 & 96.1 & 86.8 & 72.1 & \textbf{87.9} & +0.9 \\
        & CLIP-LoRA~\citep{zanella2024low} & 38.8 & 83.5 & 64.0 & 72.8 & 77.4 & 67.3 & - & 90.6 & 81.1 & 92.9 & 95.0 & 82.6 & 71.4 & 85.6 & - \\
        & \quad +FMA~\citep{jiang2025exploring}   & 40.3 & 85.0 & 67.0 & 73.7 & 78.9 & 69.0 & +1.7 & 90.8 & 82.4 & 95.0 & 95.8 & 83.2 & 72.0 & 86.5 & +0.9 \\
        & \quad +DP-FM  & 41.7 & 89.3 & 69.2 & 73.4 & 80.5 & 70.8 & \textbf{+3.5} & 92.2 & 82.3 & 96.8 & 96.0 & 83.7 & 72.1 & 87.2 & \textbf{+1.6} \\
        \midrule

        \multirow{5}{*}{\textbf{16}} 
        & CLIP-LoRA$^\dagger$~\citep{desai2023hyperbolic} & 57.6 & 91.3 & 73.7 & 77.1 & 86.5 & 77.2 & - & 93.6 & 87.3 & 98.7 & 96.4 & 87.3 & 73.6 & 89.5 & - \\
        & \quad +HFM~\citep{li2026path}   & 61.4 & 93.1 & 75.2 & 77.6 & 88.6 & \textbf{79.2} & +1.9 & 94.0 & 88.9 & 98.7 & 96.8 & 87.3 & 73.7 & \textbf{89.9} & +0.4 \\
        & CLIP-LoRA~\citep{zanella2024low} & 54.7 & 90.7 & 73.0 & 76.0 & 86.0 & 76.1 & - & 91.6 & 86.2 & 97.9 & 96.1 & 84.2 & 73.4 & 88.2 & - \\
        & \quad +FMA~\citep{jiang2025exploring}   & 57.8 & 91.0 & 75.4 & 77.2 & 87.7 & 77.8 & +1.7 & 91.6 & 87.1 & 99.1 & 96.5 & 85.1 & 73.5 & 88.8 & +0.6 \\
        & \quad +DP-FM  & 59.2 & 93.1 & 77.1 & 77.6 & 88.8 & \textbf{79.2} & \textbf{+3.1} & 92.8 & 88.0 & 99.0 & 96.6 & 85.8 & 74.4 & 89.4 & \textbf{+1.2} \\
        \midrule

        \multirow{5}{*}{\textbf{16}} 
        & CLIP-Adapter$^\dagger$~\citep{desai2023hyperbolic} & 34.2 & 72.3 & 64.1 & 74.2 & 73.6 & 63.7 & - & 92.7 & 82.4 & 93.3 & 95.1 & 87.0 & 71.2 & 87.0 & - \\
        & \quad +HFM~\citep{li2026path}   & 43.8 & 85.7 & 71.4 & 76.1 & 80.6 & 71.5 & +7.8 & 93.3 & 84.6 & 97.2 & 96.2 & 87.1 & 72.1 & 88.4 & +1.5 \\
        & CLIP-Adapter~\citep{gao2024clip} & 33.8 & 70.4 & 59.3 & 74.3 & 74.2 & 62.4 & - & 92.4 & 80.1 & 93.6 & 94.9 & 87.1 & 71.6 & 86.6 & - \\
        & \quad +FMA~\citep{jiang2025exploring}   & 35.8 & 85.6 & 69.2 & 74.4 & 74.7 & 67.9 & +5.5 & 92.9 & 81.5 & 95.6 & 96.0 & 87.2 & 71.3 & 87.4 & +0.8 \\
        & \quad +DP-FM  & 45.1 & 86.4 & 71.6 & 75.8 & 82.7 & \textbf{72.3} & \textbf{+9.9} & 92.9 & 84.6 & 97.9 & 96.0 & 87.1 & 72.4 & \textbf{88.5} & \textbf{+1.9} \\
        \bottomrule
    \end{tabular}
    }
    \vspace{-2em}
\end{table*}

\textbf{Datasets and Baselines.} We evaluated DP-FM on few-shot image
classification tasks. Specifically, we conducted experiments on
11 benchmarks, including Aircraft~\citep{maji2013fine}, EuroSAT~\citep{helber2019eurosat}, DTD~\citep{cimpoi2014describing},
SUN397~\citep{xiao2010sun}, StanfordCars~\citep{krause20133d}, OxfordPets~\citep{parkhi2012cats}, UCF101~\citep{soomro2012ucf101}, Flowers102~\citep{nilsback2008automated}, Caltech101~\citep{fei2004learning}, Food101~\citep{bossard2014food}, and ImageNet~\citep{deng2009imagenet}. Following the FMA setting, we partitioned them into two subsets: five datasets form the difficult group, and the remaining six constitute the easy group. For each dataset, we adopted the standard train/validation/test splits.
Under the K-shot setting, we constructed the training set by randomly sampling K labeled images per class, with the rest used for validation and testing. For PEFT methods, we adopted CoOp~\citep{zhou2022learning}, CoCoOp~\citep{zhou2022conditional}, TIP-Adapter~\citep{zhang2022tip}, CLIP-Adapter~\citep{gao2024clip}, PLOT++~\citep{chen2022plot}, KgCoOp~\citep{yao2023visual}, ProGrad~\citep{zhu2023prompt}, and CLIP-LoRA~\citep{zanella2024low} as our baselines. 
For multi-step FM methods, we adopted FMA~\citep{jiang2025exploring} and HFM~\citep{li2026path} as our baselines.

\noindent\textbf{Settings.} We implemented DP-FM based on the pre-trained CLIP, CLIP-LoRA~\citep{zanella2024low}, and CLIP-Adapter~\citep{gao2024clip} model. Specifically, we first extract cross-modal features from the pre-trained models, then utilize a velocity network to regress velocity fields that transport image features to text features. Following the FMA~\citep {jiang2025exploring} setting, we utilize a lightweight MAR~\citep{li2024autoregressive} network as our velocity network. In the DP-FM algorithm, we set $\text{CONSTANT}=25$, time-shift parameter $s=0.1$, dropout probability $p_{drop}=0.1$, and CFG scale $w=5.0$ across all the experiments. We trained DP-FM with AdamW~\citep{loshchilov2017decoupled} using a learning rate of $2\times10^{-4}$ and a weight decay of $0.01$. 

\noindent{\textbf{Performance against FM Baselines.}} We evaluate DP-FM against recent FM methods for few-shot adaptation, including  WP-FM (in Euclidean\footnote{WP-FM (Euclidean) is the same as FMA~\citep{jiang2025exploring}, operating on pre-trained cross-modal features}~\citep{jiang2025exploring}, Hyperbolic\footnote{To isolate the impact of different geometries, we omit HFM's centripetal hyperbolic alignment and diameter-based stopping phases, ensuring FMA, HFM, and DP-FM operate on identical pre-trained cross-modal features in 1-stage training.\label{footnote:1}}~\citep{li2026path} space). For a fair comparison, all the methods operate on the same pre-trained CLIP model and are equipped with the time-shift schedule and CFG. As shown in Table~\ref{tab:tab-1}, DP-FM consistently achieves higher average accuracy across the 1, 4, and 16-shot settings. By isolating the geometric design on the pre-trained cross-modal features, these results explicitly confirm that the decoupled cylindrical manifold provides an enhanced foundation for FM compared to flat Euclidean or Hyperbolic spaces. Furthermore, as illustrated in Figure~\ref{fig:fig-4}, DP-FM exhibits a more uniform angular speed and enhanced adaptation performance.

Furthermore, we extend our evaluation to compare DP-FM against state-of-the-art FM methods when integrated with established PEFT baselines, specifically CLIP-LoRA and CLIP-Adapter, as shown in Table \ref{tab:tab-2}. DP-FM consistently delivers the most substantial performance improvements ($\Delta$) over the baselines across different datasets. For instance, under the 16-shot setting with CLIP-Adapter, DP-FM achieves a remarkable average improvement of $+9.9\%$ on the difficult datasets, which significantly surpasses the gains provided by FMA ($+5.5\%$) and HFM ($+7.8\%$). Similarly, when built upon the CLIP-LoRA architecture, DP-FM outperforms competing FM methods in most cases, demonstrating its enhanced capability to refine feature alignment effectively. 

\begin{table*}[t]
    \centering
    \vspace{-1em}
    \caption{\textbf{Comparison with other state-of-the-art PEFT methods.} Building upon CLIP-LoRA, adaptation performance can be further improved by our DP-FM.}
    \label{tab:main_results}
    \small % 使用较小字号
    \setlength{\tabcolsep}{3.5pt} % 压缩列间距防止超宽
    \resizebox{\textwidth}{!}{
    \begin{tabular}{@{} c|l| ccccc >{\columncolor{gray!15}}l |c cccccc >{\columncolor{gray!15}}l @{}}
        \toprule
        & & \multicolumn{6}{c}{\textbf{Difficult Datasets}} & & \multicolumn{7}{c}{\textbf{Easy Datasets}} \\
        \cmidrule(lr){3-8} \cmidrule(lr){10-16}
        Shot & Method &  Aircraft & SAT & DTD & SUN & Cars & \textbf{AVG} & & Pets & UCF & Flowers & Caltech & Food & Net & \textbf{AVG} \\
        \midrule

        \multirow{1}{*}{\textbf{0}} 
        & CLIP~\citep{radford2021learning} & 24.8 & 47.8 & 43.8 & 62.6 & 65.5 & 48.9 & & 89.1 & 66.8 & 71.2 & 92.9 & 86.1 & 66.7 & 78.8 \\
        \midrule

        \multirow{10}{*}{\textbf{1}}  
        & CoOp~\citep{zhou2022learning}            & 20.8 & 56.4 & 50.1 & 67.0 & 67.5 & 52.4 & & 90.2 & 71.2 & 78.3 & 92.5 & 84.3 & 65.7 & 80.4 \\
        & CoCoOp~\citep{zhou2022conditional}          & 28.1 & 55.4 & 52.6 & 68.7 & 67.6 & 54.5 & & 91.9 & 70.4 & 73.4 & 94.1 & 84.9 & 69.4 & 80.7 \\
        & TIP-Adapter~\citep{zhang2022tip}     & 28.8 & 67.8 & 51.6 & 67.2 & 67.1 & 56.5 & & 90.6 & 73.4 & 83.8 & 94.0 & 85.8 & 69.4 & 82.8 \\
        & CLIP-Adapter~\citep{gao2024clip}    & 25.2 & 49.3 & 44.2 & 65.4 & 65.7 & 50.0 & & 89.0 & 66.9 & 71.3 & 92.0 & 86.1 & 67.9 & 78.9 \\
        & PLOT++~\citep{chen2022plot}         & 28.6 & 65.4 & 54.6 & 66.8 & 68.8 & 56.8 & & 91.9 & 74.3 & 80.5 & 94.3 & 86.2 & 66.5 & 82.3 \\
        & KgCoOp~\citep{yao2023visual}          & 26.8 & 61.9 & 52.7 & 68.4 & 66.7 & 55.3 & & 92.1 & 72.8 & 74.7 & 94.2 & 86.4 & 68.9 & 81.5 \\
        & ProGrad~\citep{zhu2023prompt}         & 28.9 & 57.0 & 52.8 & 67.0 & 68.2 & 54.8 & & 91.4 & 73.3 & 80.9 & 93.5 & 84.9 & 67.0 & 81.8 \\
        & CLIP-LoRA~\cite{zanella2024low}       & 28.0 & 71.9 & 54.1 & 70.3 & 69.4 & 58.7 & & 91.9 & 75.4 & 81.4 & 93.8 & 85.1 & 70.3 & 83.0 \\
        & \quad +DP-FM     & 29.6 & 81.7 & 56.9 & 71.0 & 70.8 & \impr{+3.3}{\textbf{62.0}} & & 93.1 & 76.8 & 89.0 & 94.4 & 85.0 & 70.3 & \impr{+1.8}{\textbf{84.8}} \\
         \midrule

        \multirow{10}{*}{\textbf{4}}  
        & CoOp~\citep{zhou2022learning}            & 30.9 & 69.7 & 59.5 & 69.7 & 74.4 & 60.8 & & 92.5 & 77.6 & 92.2 & 94.5 & 84.5 & 68.8 & 85.0 \\
        & CoCoOp~\citep{zhou2022conditional}          & 30.6 & 61.7 & 55.7 & 70.4 & 69.5 & 57.6 & & 92.7 & 75.3 & 81.5 & 94.8 & 86.3 & 70.6 & 83.5 \\
        & TIP-Adapter~\citep{zhang2022tip}     & 35.7 & 76.8 & 59.8 & 70.8 & 74.1 & 63.4 & & 91.9 & 78.1 & 92.1 & 94.8 & 86.5 & 70.7 & 85.7 \\
        & CLIP-Adapter~\citep{gao2024clip}    & 27.9 & 51.2 & 46.1 & 68.0 & 67.5 & 52.1 & & 90.8 & 70.6 & 73.1 & 94.0 & 86.5 & 68.6 & 80.6 \\
        & PLOT++~\citep{chen2022plot}          & 35.3 & 83.2 & 62.4 & 71.7 & 76.3 & 65.8 & & 92.7 & 79.8 & 92.9 & 95.1 & 86.5 & 70.4 & 86.2 \\
        & KgCoOp~\citep{yao2023visual}          & 32.2 & 71.8 & 58.7 & 71.5 & 69.5 & 60.7 & & 92.6 & 77.6 & 87.0 & 95.0 & 86.9 & 69.9 & 84.8 \\
        & ProGrad~\citep{zhu2023prompt}         & 34.1 & 69.6 & 59.7 & 71.7 & 75.0 & 62.0 & & 92.1 & 77.9 & 91.1 & 94.4 & 85.4 & 70.2 & 85.2 \\
        & CLIP-LoRA~\citep{zanella2024low}       & 38.8 & 83.5 & 64.0 & 72.8 & 77.4 & 67.3 & & 90.6 & 81.1 & 92.9 & 95.0 & 82.6 & 71.4 & 85.6 \\
        & \quad +DP-FM     & 41.7 & 89.3 & 69.2 & 73.4 & 80.5 & \impr{+3.5}{\textbf{70.8}} & & 92.2 & 82.3 & 96.8 & 96.0 & 83.7 & 72.1 & \impr{+1.6}{\textbf{87.2}} \\
        \midrule

        \multirow{10}{*}{\textbf{16}} 
        & CoOp~\citep{zhou2022learning}            & 43.3 & 86.0 & 70.0 & 74.9 & 83.1 & 71.5 & & 91.1 & 83.1 & 97.2 & 95.5 & 84.4 & 71.4 & 87.1 \\
        & CoCoOp~\citep{zhou2022conditional}          & 33.8 & 75.5 & 65.8 & 72.8 & 72.4 & 64.1 & & 93.2 & 76.0 & 87.1 & 95.2 & 87.4 & 71.1  & 85.0 \\
        & TIP-Adapter~\citep{zhang2022tip}     & 44.6 & 85.9 & 70.8 & 76.0 & 82.3 & 71.9 & & 92.6 & 83.9 & 96.2 & 95.7 & 86.8 & 73.4 & 88.1 \\
        & CLIP-Adapter~\citep{gao2024clip}    & 34.2 & 71.4 & 59.4 & 74.2 & 74.0 & 62.6 & & 92.3 & 80.2 & 92.9 & 94.9 & 87.1 & 69.8 & 86.2 \\
        & PLOT++~\citep{chen2022plot}          & 46.7 & 92.0 & 71.4 & 76.0 & 84.6 & 74.1 & & 93.6 & 85.3 & 97.6 & 96.0 & 87.1 & 72.6 & 88.7 \\
        & KgCoOp~\citep{yao2023visual}          & 36.5 & 76.2 & 68.7 & 73.3 & 74.8 & 65.9 & & 93.2 & 81.7 & 93.4 & 95.2 & 87.2 & 70.4 & 86.9 \\
        & ProGrad~\citep{zhu2023prompt}         & 43.0 & 83.6 & 68.8 & 75.1 & 82.9 & 70.7 & & 92.8 & 82.7 & 96.6 & 95.9 & 85.8 & 72.1 & 87.7 \\
        & CLIP-LoRA~\citep{zanella2024low}       & 54.7 & 90.7 & 73.0 & 76.0 & 86.0 & 76.1 & & 91.6 & 86.2 & 97.9 & 96.1 & 84.2 & 73.4 & 88.2 \\
        & \quad +DP-FM     & 59.2 & 93.1 & 77.1 & 77.6 & 88.8 & \impr{+3.1}{\textbf{79.2}} & & 92.8 & 88.0 & 99.0 & 96.6 & 85.8 & 74.4 & \impr{+1.2}{\textbf{89.4}} \\
        \midrule
    \end{tabular}
    }
    \vspace{-0.5em}
\end{table*}

\begin{table*}[t]
    \centering
    \vspace{-1em}
    \caption{\textbf{Ablation study} on Time-shift Schedule (\textbf{T}), Radius component (\textbf{R}), and Classifier-free guidance (\textbf{C}) under the 16-shot setting. The numbers in parentheses denote CFG scale $w$.}
    \label{tab:ablation_full}
    \small
    \setlength{\tabcolsep}{3.5pt}  
    \resizebox{\textwidth}{!}{
    \begin{tabular}{@{} c | c c c | ccccc >{\columncolor{gray!15}}l | cccccc >{\columncolor{gray!15}}l @{}}
        \toprule
        Method & T & R & C &  Aircraft & SAT & DTD & SUN & Cars & \textbf{AVG} & Pets & UCF & Flowers & Caltech & Food & Net & \textbf{AVG} \\
        \midrule
        CLIP~\citep{radford2021learning}  & $\times$ & $\times$ & $\times$ & 24.8 & 47.8 & 43.8 & 62.6 & 65.5 & 48.9 & 89.1 & 66.8 & 71.2 & 92.9 & 86.1 & 66.7 & 78.8 \\
        \midrule
        \multirow{6}{*}{+DP-FM} 
        & $\times$ & $\times$ & $\times$ & 42.7 & 85.5 & 70.8 & 73.7 & 82.9 & \impr{+22.2}{71.1} & 92.9 & 82.3 & 97.3 & 95.9 & 86.8 & 70.8 & \impr{+8.9}{87.7} \\
        & $\times$   & \checkmark & \checkmark (5) & 41.3 & 86.7 & 69.8 & 72.4 & 79.8 & \impr{+21.1}{70.0} & 92.1 & 82.6 & 97.3 & 95.2 & 86.4 & 71.0 & \impr{+8.6}{87.4} \\
        & \checkmark & $\times$   & \checkmark (5) & 44.2 & 86.8 & 69.6 & 73.6 & 83.6 & \impr{+22.7}{71.6} & 93.0 & 82.6 & 97.8 & 96.1 & 86.9 & 71.0 & \impr{+9.1}{87.9} \\
        & \checkmark & \checkmark & $\times$           & 44.6 & 87.0 & 71.1 & 73.8 & 82.6 & \impr{+22.9}{71.8} & 92.7 & 82.6 & 97.7 & 95.9 & 87.1 & 71.1 & \impr{+9.1}{87.9} \\
        & \checkmark & \checkmark & \checkmark (1) & 45.8 & 87.8 & 71.2 & 73.1 & 82.6 & \impr{+23.2}{72.1} & 93.1 & 82.6 & 97.7 & 96.2 & 87.0 & 71.1 & \impr{+9.2}{\textbf{88.0}} \\
        & \checkmark & \checkmark & \checkmark (5) & 46.0 & 87.8 & 71.2 & 73.8 & 83.6 & \impr{+23.6}{\textbf{72.5}} & 93.2 & 83.4 & 97.8 & 96.2 & 87.0 & 71.1 & \impr{+9.3}{\textbf{88.1}} \\
        
        \bottomrule
    \end{tabular}
    }
    \vspace{-1.5em}
\end{table*}

\noindent{\textbf{Comparison with state-of-the-art PEFT.}} As presented in Table \ref{tab:main_results}, we evaluate the performance of our DP-FM against a comprehensive set of single-step PEFT methods. Built upon CLIP-LoRA, DP-FM consistently establishes new state-of-the-art results across 1, 4, and 16-shot settings for both difficult and easy dataset groups. Most notably, under the 16-shot setting, DP-FM achieves an average accuracy of $79.2\%$ on the difficult datasets and $89.4\%$ on the easy datasets. This performance significantly exceeds strong single-step baselines such as PLOT++ ($74.1\%$ and $88.7\%$) and the standalone CLIP-LoRA ($76.1\%$ and $88.2\%$). These results confirm that modeling the adaptation process as a continuous flow on a decoupled cylindrical manifold yields superior generalization compared to conventional static, single-step prompting or adaptation strategies. 

\noindent{\textbf{Ablation Study and Component Analysis.}} To analyze the contribution of each proposed component, we conduct ablation experiments under the $16$-shot setting (Table~\ref{tab:ablation_full}). First, we examine the effect of the Time-shift Schedule (T) on training convergence. Notably, adding the Radius component (R) and classifier-free guidance (C) without T (Row 3) yields a lower performance ($70.0\%$) than the baseline lacking all three components (Row 2, $71.1\%$). This is likely because introducing extra radial and conditional information increases optimization complexity, which delays the training process. The time-shift schedule addresses this by prioritizing training of early-stage velocity estimation. With T stabilizing the optimization, the empirical benefits of R and C become clearly evident, as integrating the radial component and CFG both lead to consistent performance improvements. Ultimately, combining all three components achieves the highest average accuracy of $72.5\%$ on difficult datasets, validating the integrated design of our method.

% With T stabilizing the optimization, the individual contributions of R and C can be observed: preserving radial dynamics helps maintain modality confidence, while CFG incorporates dataset-specific information. Ultimately, combining all three components achieves the highest average accuracy of $72.5\%$ on difficult datasets among the tested configurations, supporting the integrated design of DP-FM.

% \vspace{-0.5em}
% \section{Conclusion}
% \vspace{-0.5em}
% In this paper, we investigated the fundamental geometric constraints of existing FM methods for the few-shot adaptation of VLMs. From a polar decomposition perspective, we demonstrated that previous FM methods inherently suffer from angular dynamics distortion, radial dynamics neglect, and context-agnostic unconditional flows. To address these limitations, we introduced warped product flow matching,\ie, WP-FM, a unified Riemannian framework. By adopting an optimal constant-warping metric, we derived direct product flow matching: DP-FM, which strictly decouples radial and angular dynamics. This formulation ensures constant-speed angular geodesic transport while explicitly preserving modality confidence. Furthermore, we integrated classifier-free guidance conditioned on the pre-trained model's hidden states to recover dataset-specific information. Extensive experiments across 11 diverse few-shot vision-language benchmarks demonstrate that our method achieves enhanced adaptation accuracy during multi-step alignment. 
\vspace{-0.5em}
\section{Conclusion}

In this paper, we investigated the geometric limitations of existing FM methods for few-shot VLM adaptation, demonstrating that they inherently suffer from angular dynamics distortion, radial dynamics neglect, and context-agnostic flows. To address this, we introduced warped product flow matching (WP-FM), a unified Riemannian framework. By adopting a constant-warping metric, we derived DP-FM to strictly decouple radial and angular dynamics, ensuring constant-speed geodesic transport and preserving modality confidence. Additionally, we integrated hidden-state conditioned classifier-free guidance to recover dataset-specific context. Extensive experiments across 11 vision-language benchmarks confirm our method achieves enhanced adaptation accuracy during multi-step alignment.

\clearpage
\bibliography{neurips_2026}

@inproceedings{radford2021learning,
  title={Learning transferable visual models from natural language supervision},
  author={Radford, Alec and Kim, Jong Wook and Hallacy, Chris and Ramesh, Aditya and Goh, Gabriel and Agarwal, Sandhini and Sastry, Girish and Askell, Amanda and Mishkin, Pamela and Clark, Jack and others},
  booktitle={ICML},
  pages={8748--8763},
  year={2021},
  organization={PMLR}
}

@article{alayrac2022flamingo,
  title={Flamingo: a visual language model for few-shot learning},
  author={Alayrac, Jean-Baptiste and Donahue, Jeff and Luc, Pauline and Miech, Antoine and Barr, Iain and Hasson, Yana and Lenc, Karel and Mensch, Arthur and Millican, Katherine and Reynolds, Malcolm and others},
  journal={NeurIPS},
  volume={35},
  pages={23716--23736},
  year={2022}
}

@inproceedings{li2023blip,
  title={Blip-2: Bootstrapping language-image pre-training with frozen image encoders and large language models},
  author={Li, Junnan and Li, Dongxu and Savarese, Silvio and Hoi, Steven},
  booktitle={ICML},
  pages={19730--19742},
  year={2023},
  organization={PMLR}
}

@article{liu2023visual,
  title={Visual instruction tuning},
  author={Liu, Haotian and Li, Chunyuan and Wu, Qingyang and Lee, Yong Jae},
  journal={NeurIPS},
  volume={36},
  pages={34892--34916},
  year={2023}
}

@inproceedings{li2022blip,
  title={Blip: Bootstrapping language-image pre-training for unified vision-language understanding and generation},
  author={Li, Junnan and Li, Dongxu and Xiong, Caiming and Hoi, Steven},
  booktitle={ICML},
  pages={12888--12900},
  year={2022},
  organization={PMLR}
}

@article{hu2022lora,
  title={Lora: Low-rank adaptation of large language models.},
  author={Hu, Edward J and Shen, Yelong and Wallis, Phillip and Allen-Zhu, Zeyuan and Li, Yuanzhi and Wang, Shean and Wang, Liang and Chen, Weizhu and others},
  journal={ICLR},
  volume={1},
  number={2},
  pages={3},
  year={2022}
}

@article{zhou2022learning,
  title={Learning to prompt for vision-language models},
  author={Zhou, Kaiyang and Yang, Jingkang and Loy, Chen Change and Liu, Ziwei},
  journal={IJCV},
  volume={130},
  number={9},
  pages={2337--2348},
  year={2022},
  publisher={Springer}
}

@inproceedings{zhou2022conditional,
  title={Conditional prompt learning for vision-language models},
  author={Zhou, Kaiyang and Yang, Jingkang and Loy, Chen Change and Liu, Ziwei},
  booktitle={CVPR},
  pages={16816--16825},
  year={2022}
}

@inproceedings{zanella2024low,
  title={Low-rank few-shot adaptation of vision-language models},
  author={Zanella, Maxime and Ben Ayed, Ismail},
  booktitle={Proceedings of the IEEE/CVF Conference on Computer Vision and Pattern Recognition},
  pages={1593--1603},
  year={2024}
}

@inproceedings{jiang2025exploring,
  title={Exploring Cross-Modal Flows for Few-Shot Learning},
  author={Jiang, Ziqi and Wang, Yanghao and Chen, Long},
  booktitle={ICLR},
  year={2026}
}

@article{lipman2022flow,
  title={Flow matching for generative modeling},
  author={Lipman, Yaron and Chen, Ricky TQ and Ben-Hamu, Heli and Nickel, Maximilian and Le, Matt},
  journal={arXiv preprint},
  year={2022}
}

@article{liu2022flow,
  title={Flow straight and fast: Learning to generate and transfer data with rectified flow},
  author={Liu, Xingchao and Gong, Chengyue and Liu, Qiang},
  journal={arXiv preprint},
  year={2022}
}

@article{li2026path,
  title={Path-Decoupled Hyperbolic Flow Matching for Few-Shot Adaptation},
  author={Li, Lin and Jiang, Ziqi and Ye, Gefan and He, Zhenqi and Li, Jiahui and Xiao, Jun and Cheng, Kwang-Ting and Chen, Long},
  journal={arXiv preprint},
  year={2026}
}

@inproceedings{jo2026angular,
  title={Angular Gradient Sign Method: Uncovering Vulnerabilities in Hyperbolic Networks},
  author={Jo, Minsoo and Yang, Dongyoon and Kim, Taesup},
  booktitle={AAAI},
  volume={40},
  number={7},
  pages={5566--5574},
  year={2026}
}

@article{chen2023flow,
  title={Flow matching on general geometries},
  author={Chen, Ricky TQ and Lipman, Yaron},
  journal={arXiv preprint},
  year={2023}
}

@inproceedings{esser2024scaling,
  title={Scaling rectified flow transformers for high-resolution image synthesis},
  author={Esser, Patrick and Kulal, Sumith and Blattmann, Andreas and Entezari, Rahim and M{\"u}ller, Jonas and Saini, Harry and Levi, Yam and Lorenz, Dominik and Sauer, Axel and Boesel, Frederic and others},
  booktitle={ICML},
  year={2024}
}

@book{o1983semi,
  title={Semi-Riemannian geometry with applications to relativity},
  author={O'neill, Barrett},
  volume={103},
  year={1983},
  publisher={Academic press}
}

@inproceedings{zhai2023sigmoid,
  title={Sigmoid loss for language image pre-training},
  author={Zhai, Xiaohua and Mustafa, Basil and Kolesnikov, Alexander and Beyer, Lucas},
  booktitle={ICCV},
  pages={11975--11986},
  year={2023}
}

@article{han2024parameter,
  title={Parameter-efficient fine-tuning for large models: A comprehensive survey},
  author={Han, Zeyu and Gao, Chao and Liu, Jinyang and Zhang, Jeff and Zhang, Sai Qian},
  journal={arXiv preprint},
  year={2024}
}

@inproceedings{jia2021scaling,
  title={Scaling up visual and vision-language representation learning with noisy text supervision},
  author={Jia, Chao and Yang, Yinfei and Xia, Ye and Chen, Yi-Ting and Parekh, Zarana and Pham, Hieu and Le, Quoc and Sung, Yun-Hsuan and Li, Zhen and Duerig, Tom},
  booktitle={ICML},
  pages={4904--4916},
  year={2021},
  organization={PMLR}
}

@inproceedings{houlsby2019parameter,
  title={Parameter-efficient transfer learning for NLP},
  author={Houlsby, Neil and Giurgiu, Andrei and Jastrzebski, Stanislaw and Morrone, Bruna and De Laroussilhe, Quentin and Gesmundo, Andrea and Attariyan, Mona and Gelly, Sylvain},
  booktitle={ICML},
  pages={2790--2799},
  year={2019},
  organization={PMLR}
}

@article{gao2024clip,
  title={Clip-adapter: Better vision-language models with feature adapters},
  author={Gao, Peng and Geng, Shijie and Zhang, Renrui and Ma, Teli and Fang, Rongyao and Zhang, Yongfeng and Li, Hongsheng and Qiao, Yu},
  journal={IJCV},
  volume={132},
  number={2},
  pages={581--595},
  year={2024},
  publisher={Springer}
}

@article{chen2026bi,
  title={Bi-Anchor Interpolation Solver for Accelerating Generative Modeling},
  author={Chen, Hongxu and Li, Hongxiang and Wang, Zhen and Chen, Long},
  journal={arXiv preprint},
  year={2026}
}

@inproceedings{ma2024sit,
  title={Sit: Exploring flow and diffusion-based generative models with scalable interpolant transformers},
  author={Ma, Nanye and Goldstein, Mark and Albergo, Michael S and Boffi, Nicholas M and Vanden-Eijnden, Eric and Xie, Saining},
  booktitle={ECCV},
  pages={23--40},
  year={2024},
  organization={Springer}
}

@inproceedings{peebles2023scalable,
  title={Scalable diffusion models with transformers},
  author={Peebles, William and Xie, Saining},
  booktitle={ICCV},
  pages={4195--4205},
  year={2023}
}

@inproceedings{rombach2022high,
  title={High-resolution image synthesis with latent diffusion models},
  author={Rombach, Robin and Blattmann, Andreas and Lorenz, Dominik and Esser, Patrick and Ommer, Bj{\"o}rn},
  booktitle={CVPR},
  pages={10684--10695},
  year={2022}
}

@article{song2020score,
  title={Score-based generative modeling through stochastic differential equations},
  author={Song, Yang and Sohl-Dickstein, Jascha and Kingma, Diederik P and Kumar, Abhishek and Ermon, Stefano and Poole, Ben},
  journal={arXiv preprint},
  year={2020}
}

@article{ho2020denoising,
  title={Denoising diffusion probabilistic models},
  author={Ho, Jonathan and Jain, Ajay and Abbeel, Pieter},
  journal={NeurIPS},
  volume={33},
  pages={6840--6851},
  year={2020}
}

@article{teng2023relay,
  title={Relay diffusion: Unifying diffusion process across resolutions for image synthesis},
  author={Teng, Jiayan and Zheng, Wendi and Ding, Ming and Hong, Wenyi and Wangni, Jianqiao and Yang, Zhuoyi and Tang, Jie},
  journal={arXiv preprint},
  year={2023}
}

@inproceedings{hoogeboom2023simple,
  title={simple diffusion: End-to-end diffusion for high resolution images},
  author={Hoogeboom, Emiel and Heek, Jonathan and Salimans, Tim},
  booktitle={ICML},
  pages={13213--13232},
  year={2023},
  organization={PMLR}
}

@article{ho2022classifier,
  title={Classifier-free diffusion guidance},
  author={Ho, Jonathan and Salimans, Tim},
  journal={arXiv preprint},
  year={2022}
}

@article{maji2013fine,
  title={Fine-grained visual classification of aircraft},
  author={Maji, Subhransu and Rahtu, Esa and Kannala, Juho and Blaschko, Matthew and Vedaldi, Andrea},
  journal={arXiv preprint},
  year={2013}
}

@article{helber2019eurosat,
  title={Eurosat: A novel dataset and deep learning benchmark for land use and land cover classification},
  author={Helber, Patrick and Bischke, Benjamin and Dengel, Andreas and Borth, Damian},
  journal={J-STARS},
  volume={12},
  number={7},
  pages={2217--2226},
  year={2019},
  publisher={IEEE}
}

@inproceedings{cimpoi2014describing,
  title={Describing textures in the wild},
  author={Cimpoi, Mircea and Maji, Subhransu and Kokkinos, Iasonas and Mohamed, Sammy and Vedaldi, Andrea},
  booktitle={CVPR},
  pages={3606--3613},
  year={2014}
}

@inproceedings{xiao2010sun,
  title={Sun database: Large-scale scene recognition from abbey to zoo},
  author={Xiao, Jianxiong and Hays, James and Ehinger, Krista A and Oliva, Aude and Torralba, Antonio},
  booktitle={CVPR},
  pages={3485--3492},
  year={2010},
  organization={IEEE}
}

@inproceedings{krause20133d,
  title={3d object representations for fine-grained categorization},
  author={Krause, Jonathan and Stark, Michael and Deng, Jia and Fei-Fei, Li},
  booktitle={ICCV},
  pages={554--561},
  year={2013}
}

@inproceedings{parkhi2012cats,
  title={Cats and dogs},
  author={Parkhi, Omkar M and Vedaldi, Andrea and Zisserman, Andrew and Jawahar, CV},
  booktitle={CVPR},
  pages={3498--3505},
  year={2012},
  organization={IEEE}
}

@article{soomro2012ucf101,
  title={Ucf101: A dataset of 101 human actions classes from videos in the wild},
  author={Soomro, Khurram and Zamir, Amir Roshan and Shah, Mubarak},
  journal={arXiv preprint},
  year={2012}
}

@inproceedings{nilsback2008automated,
  title={Automated flower classification over a large number of classes},
  author={Nilsback, Maria-Elena and Zisserman, Andrew},
  booktitle={ICVGIP},
  pages={722--729},
  year={2008},
  organization={IEEE}
}

@inproceedings{fei2004learning,
  title={Learning generative visual models from few training examples: An incremental bayesian approach tested on 101 object categories},
  author={Fei-Fei, Li and Fergus, Rob and Perona, Pietro},
  booktitle={CVPR},
  pages={178--178},
  year={2004},
  organization={IEEE}
}

@inproceedings{bossard2014food,
  title={Food-101--mining discriminative components with random forests},
  author={Bossard, Lukas and Guillaumin, Matthieu and Van Gool, Luc},
  booktitle={ECCV},
  pages={446--461},
  year={2014},
  organization={Springer}
}

@inproceedings{deng2009imagenet,
  title={Imagenet: A large-scale hierarchical image database},
  author={Deng, Jia and Dong, Wei and Socher, Richard and Li, Li-Jia and Li, Kai and Fei-Fei, Li},
  booktitle={CVPR},
  pages={248--255},
  year={2009},
  organization={Ieee}
}

@article{li2024autoregressive,
  title={Autoregressive image generation without vector quantization},
  author={Li, Tianhong and Tian, Yonglong and Li, He and Deng, Mingyang and He, Kaiming},
  journal={NeurIPS},
  volume={37},
  pages={56424--56445},
  year={2024}
}

@article{loshchilov2017decoupled,
  title={Decoupled weight decay regularization},
  author={Loshchilov, Ilya and Hutter, Frank},
  journal={arXiv preprint},
  year={2017}
}

@inproceedings{zhang2022tip,
  title={Tip-adapter: Training-free adaption of clip for few-shot classification},
  author={Zhang, Renrui and Zhang, Wei and Fang, Rongyao and Gao, Peng and Li, Kunchang and Dai, Jifeng and Qiao, Yu and Li, Hongsheng},
  booktitle={ECCV},
  pages={493--510},
  year={2022},
  organization={Springer}
}

@article{chen2022plot,
  title={Plot: Prompt learning with optimal transport for vision-language models},
  author={Chen, Guangyi and Yao, Weiran and Song, Xiangchen and Li, Xinyue and Rao, Yongming and Zhang, Kun},
  journal={arXiv preprint},
  year={2022}
}

@inproceedings{yao2023visual,
  title={Visual-language prompt tuning with knowledge-guided context optimization},
  author={Yao, Hantao and Zhang, Rui and Xu, Changsheng},
  booktitle={CVPR},
  pages={6757--6767},
  year={2023}
}

@inproceedings{zhu2023prompt,
  title={Prompt-aligned gradient for prompt tuning},
  author={Zhu, Beier and Niu, Yulei and Han, Yucheng and Wu, Yue and Zhang, Hanwang},
  booktitle={ICCV},
  pages={15659--15669},
  year={2023}
}

@inproceedings{rao2022denseclip,
  title={Denseclip: Language-guided dense prediction with context-aware prompting},
  author={Rao, Yongming and Zhao, Wenliang and Chen, Guangyi and Tang, Yansong and Zhu, Zheng and Huang, Guan and Zhou, Jie and Lu, Jiwen},
  booktitle={CVPR},
  pages={18082--18091},
  year={2022}
}

@inproceedings{park2023understanding,
  title={Understanding the feature norm for out-of-distribution detection},
  author={Park, Jaewoo and Chai, Jacky Chen Long and Yoon, Jaeho and Teoh, Andrew Beng Jin},
  booktitle={ICCV},
  pages={1557--1567},
  year={2023}
}

@inproceedings{desai2023hyperbolic,
  title={Hyperbolic image-text representations},
  author={Desai, Karan and Nickel, Maximilian and Rajpurohit, Tanmay and Johnson, Justin and Vedantam, Shanmukha Ramakrishna},
  booktitle={ICML},
  pages={7694--7731},
  year={2023},
  organization={PMLR}
}
\bibliographystyle{unsrt}
% \bibliography{main}
\clearpage
\appendix

%%%%%%%%%%%%%%%%%%%%%%%%%%%%%%%%%%%%%%%%%%%%%%%%%%%%%%%%%%%%
\section{Angular Truncation Error of WP-FM}
\label{app:rigorous_proof}

In this section, we provide a detailed derivation of the angular truncation error for first-order Riemannian ODE solvers under the warped product geometry introduced in Sec.~3.

\vspace{0.3em}
\noindent
\textbf{Proposition 1 (Angular Truncation Error).}
Let $(\mathcal{M}, g_\phi)$ be the warped product manifold defined in Sec.~3.1:
\begin{equation}
\mathcal{M} = \mathcal{M}_r \times_{g_\phi} \mathcal{M}_{\bm{\theta}}, 
\quad 
g_\phi = dr^2 + \phi(r)^2 d{\bm{\theta}}^2,
\end{equation}
where $\mathcal{M}_r = \mathbb{R}_{>0}$, $\mathcal{M}_{\bm{\theta}} = \mathbb{S}^{d-1}$, $\phi \in C^2(\mathbb{R}_{>0})$ and $\phi(r) > 0$. Let $\gamma(t) = (\gamma_r(t), \gamma_{\bm{\theta}}(t))$ be a unit-speed geodesic defined on $t \in [0,T]$. Denote the angular velocity magnitude (speed) as $\omega(t) := \|\dot{\gamma}_{\bm{\theta}}(t)\|_2$. Consider the first-order Riemannian integrator applied to the angular component:
\begin{equation}
\hat{\gamma}_{\bm{\theta}}(t+\Delta t) 
= \operatorname{Exp}_{\gamma_{\bm{\theta}}(t)}^{\mathcal{M}_{\bm{\theta}}} 
\big( \Delta t\, \dot{\gamma}_{\bm{\theta}}(t) \big).
\end{equation}
Then there exists a constant $C > 0$, depending only on bounded trajectory quantities, such that the angular truncation error $\epsilon_{\bm{\theta}}$ satisfies:
\begin{equation}
\epsilon_{\bm{\theta}}
:= d_{\mathcal{M}_{\bm{\theta}}}\big(\hat{\gamma}_{\bm{\theta}}(t+\Delta t), \gamma_{\bm{\theta}}(t+\Delta t)\big)
\le C \, \Delta t^2 
\left|
\frac{\phi'(\gamma_r(t))}{\phi(\gamma_r(t))} \dot{\gamma}_r(t)
\right|
\|\dot{\gamma}_{\bm{\theta}}(t)\|_2
+ O(\Delta t^3).
\end{equation}
Moreover, under the non-degeneracy condition $\dot{\gamma}_r(t) \neq 0$ and $\dot{\gamma}_{\bm{\theta}}(t) \neq \mathbf{0}$, the leading-order term vanishes if and only if:
\begin{equation}
\phi'(r) \equiv 0.
\end{equation}

\vspace{0.5em}
\noindent
\textbf{Proof.}

\vspace{0.3em}
\noindent
\textbf{Local truncation error on Riemannian manifolds.} \textit{Lemma 1 (Exponential Euler local error).} Let $(\mathcal{N}, h)$ be a Riemannian manifold and $x(t)$ a $C^2$ curve. Then:
\begin{equation}
d_{\mathcal{N}}\big(\operatorname{Exp}_{x(t)}(\Delta t\, \dot{x}(t)),\, x(t+\Delta t)\big)
= \frac{1}{2}\Delta t^2 \|\nabla_{\dot{x}} \dot{x}\|_{h}
+ O(\Delta t^3),
\end{equation}
where the remainder is uniform on compact intervals.

\vspace{0.3em}
\noindent
\textbf{Levi-Civita connection under warped product geometry.} For $\gamma(t) = (\gamma_r(t), \gamma_{\bm{\theta}}(t))$, the Levi-Civita connection under metric $g_\phi$ satisfies:
\begin{equation}
\nabla_{\dot{\gamma}} \dot{\gamma}
=
\Big(
\ddot{\gamma}_r - \phi(\gamma_r)\phi'(\gamma_r)\|\dot{\gamma}_{\bm{\theta}}\|_2^2,\;
\frac{D}{dt}\dot{\gamma}_{\bm{\theta}} 
+ 2 \frac{\phi'(\gamma_r)}{\phi(\gamma_r)} \dot{\gamma}_r \dot{\gamma}_{\bm{\theta}}
\Big).
\end{equation}

Since $\gamma$ is a geodesic on $\mathcal{M}$, we have:
\begin{equation}
\nabla_{\dot{\gamma}} \dot{\gamma} = \mathbf{0},
\end{equation}
which directly implies the angular acceleration component is strictly zero:
\begin{equation}
\frac{D}{dt}\dot{\gamma}_{\bm{\theta}} 
= -2 \frac{\phi'(\gamma_r)}{\phi(\gamma_r)} \dot{\gamma}_r \dot{\gamma}_{\bm{\theta}}.
\tag{\ref{eq4}}
\label{eq:angular_acc_appendix}
\end{equation}

\vspace{0.3em}
\noindent
\textbf{Angular truncation error.} Applying Lemma 1 on the angular sub-manifold $\mathcal{M}_{\bm{\theta}}$ gives:
\begin{equation}
\epsilon_{\bm{\theta}}
= \frac{1}{2}\Delta t^2 
\left\|
\frac{D}{dt}\dot{\gamma}_{\bm{\theta}}
\right\|_2
+ O(\Delta t^3).
\end{equation}
Substituting Eq.~\eqref{eq4}, we obtain the magnitude of the angular acceleration:
\begin{equation}
\left\|
\frac{D}{dt}\dot{\gamma}_{\bm{\theta}}
\right\|_2
= 2 \left|
\frac{\phi'(\gamma_r)}{\phi(\gamma_r)} \dot{\gamma}_r
\right|
\|\dot{\gamma}_{\bm{\theta}}\|_2.
\end{equation}

Therefore, the truncation error is:
\begin{equation}
\epsilon_{\bm{\theta}}
= \Delta t^2 
\left|
\frac{\phi'(\gamma_r)}{\phi(\gamma_r)} \dot{\gamma}_r
\right|
\|\dot{\gamma}_{\bm{\theta}}\|_2
+ O(\Delta t^3).
\end{equation}

\vspace{0.3em}
\noindent
\textbf{Characterization of vanishing error.}
The leading-order term vanishes iff:
\begin{equation}
\left|\frac{\phi'(\gamma_r)}{\phi(\gamma_r)} \dot{\gamma}_r\right| \|\dot{\gamma}_{\bm{\theta}}\|_2 = 0.
\end{equation}
Under the non-degeneracy condition $\dot{\gamma}_r \neq 0$ and $\dot{\gamma}_{\bm{\theta}} \neq \mathbf{0}$, this holds iff:
\begin{equation}
\phi'(r) \equiv 0,
\end{equation}
i.e., the warping function $\phi$ is constant.

This result characterizes the source of angular dynamics distortion in WP-FM. Any non-constant warping function introduces a coupling term proportional to $\frac{\phi'(r)}{\phi(r)}\dot{\gamma}_r$, which induces angular acceleration (Eq.~\eqref{eq4}) and leads to a second-order truncation error on the angular sub-manifold of pre-trained cross-modal features. In the Euclidean case $\phi(r)=r$, this coefficient reduces to $\frac{\dot{\gamma}_r}{\gamma_r}$, which becomes large when $\gamma_r \to 0$,  explaining the instability of angular updates observed in Sec.~3.

\begin{wrapfigure}{r}{0.6\textwidth} 
    \centering
    \includegraphics[width=\linewidth]{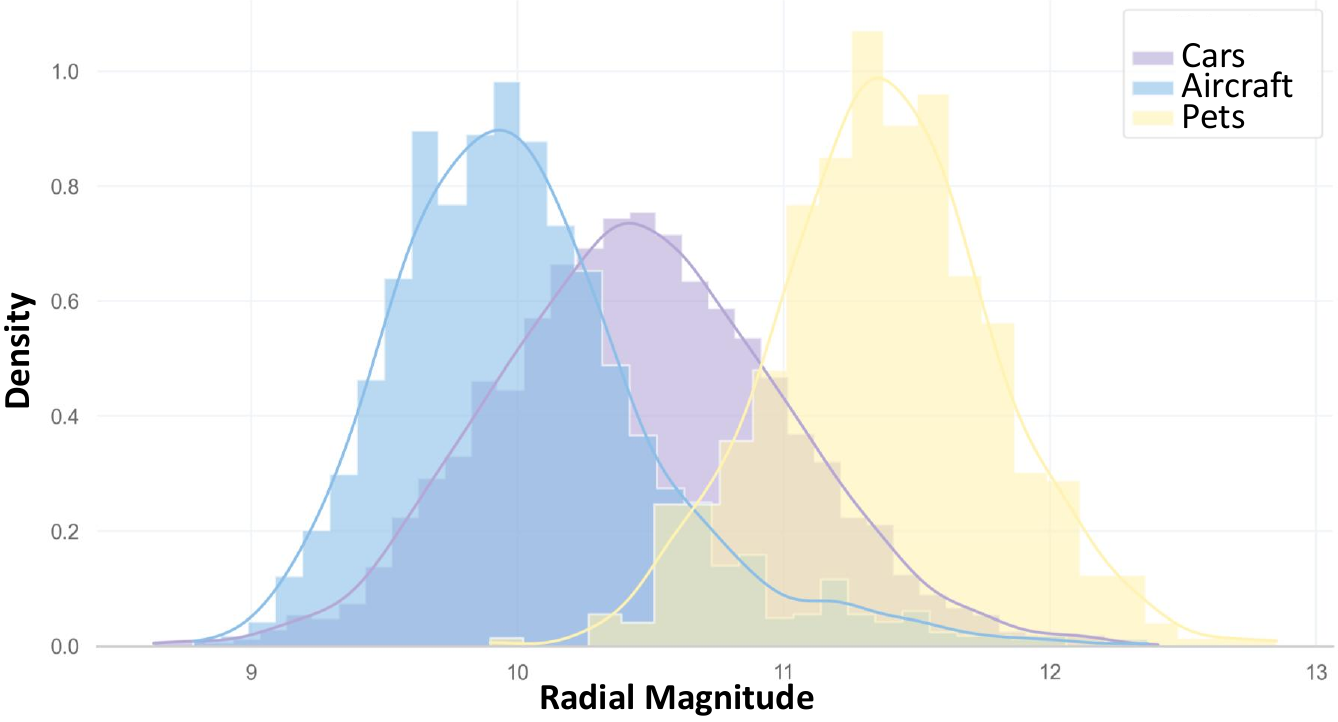}
    \caption{\textbf{Radial Magnitude Distribution across Datasets.}} 
    \label{fig:fig-5}
\end{wrapfigure}
\section{Radial Information Reflects Modality Confidence} \label{app:modality_confidence}
As illustrated in Figure~\ref{fig:fig-5}, we visualize the statistical distribution of the radial magnitudes (\ie, the $L_2$ norm) of image features extracted by the pre-trained CLIP image encoder across three different datasets. The overall magnitude distribution exhibits a clear hierarchical trend, ranking as OxfordPets $>$ StanfordCars $>$ FGVCAircraft. Notably, this trend aligns perfectly with the zero-shot classification accuracy of the pre-trained model on these respective datasets: 89.1\% for OxfordPets, 65.5\% for StanfordCars, and 24.8\% for FGVCAircraft. 

This positive correlation provides compelling empirical evidence that the radial dimension intrinsically encodes the model's modality confidence. Specifically, when the model processes images from a domain it is highly confident in (such as OxfordPets), it produces features with larger radial magnitudes. Conversely, for more challenging or out-of-distribution domains (such as FGVCAircraft), the resulting feature magnitudes are significantly smaller. Consequently, these findings emphasize that radial information serves as a crucial explicit signal for distinguishing in-distribution data from uncertain samples, and discarding it inherently blinds the velocity network, degrading the performance observed in the ablation study.

\section{Limitations}
To resolve the alignment of highly entangled cross-modal distributions, this method (DP-FM) models the alignment process as a continuous multi-step probability flow. Therefore, during the inference phase, the model relies on multi-step integration to progressively transport features, which differs from traditional parameter-efficient fine-tuning (PEFT) methods that only require a single-step adjustment. Objectively speaking, this does mean that multiple steps must be executed during runtime. However, this multi-step mechanism is precisely the core foundation that enables the Flow Matching framework to improve alignment and achieve enhanced performance.

\newpage

\end{document}